# Privacy–Personalization Trade-offs in LLMs: The Impact of Stylometric Signal Reduction on User-Specific Text Generation

Muhammed Nazmul Arefin*

*Information and Computer Science, King Fahd University of Petroleum and Minerals, Dhahran, Dammam, Saudi Arabia, g202416760@kfupm.edu.sa*

Omar Jamal Hammad

*Information and Computer Science, King Fahd University of Petroleum and Minerals, Dhahran, Dammam, Saudi Arabia, omarjh@kfupm.edu.sa*

## ABSTRACT

Large language models (LLMs) have demonstrated the ability to generate user-specific text with high stylistic fidelity. However, the personal data that enables such personalization frequently embeds demographic, cultural, and stylistic markers that raises concerns about stylometric re-identification. This paper investigates whether reducing identifiable stylistic signals affects personalization in text generation by LLMs. We introduce a controlled framework to isolate stylometric signals in LLM personalization using the LaMP-7 Twitter benchmark. Experiments on 250 sampled users compare two settings: paraphrasing conditioned on the original profile and paraphrasing conditioned on an anonymized converted profile in which demographic identifiers, cultural references, personal details, and informal linguistic cues have been systematically neutralized. Outputs are assessed by two independent LLM judges and a complementary human evaluation. Our pairwise evaluation shows that outputs conditioned on original profiles are nearly indistinguishable from human-authored ground truth, indicating that modern LLMs can closely reproduce an author's writing style with sufficient fidelity. In contrast, preference for model outputs with anonymized profiles drops to 13.0% on average, while semantic context preservation remains high at 94.8%. A study with human evaluators confirms the same pattern. These findings reveal a clear privacy-personalization trade-off and highlight the need for privacy-aware personalization methods that retain meaning while suppressing identifying stylistic signals.

## 1 Introduction

Large language models (LLMs) have become a foundational base for a rapidly expanding range of user-facing applications. This technology is now used for writing assistance, conversational agents, summarization, and creative generation. In each deployment (e.g., system, product, or solution), there is a consistent need to adapt the requirements to different users. Personalization is the ability of the model to produce outputs that reflect a specific user's communicative preferences, tone, vocabulary, and topical interests, rather than a single generic style. Retrieval-augmented generation (RAG), originally introduced by Lewis et al., is a framework for combining parametric model knowledge with non-parametric memory retrieved from an external corpus [1]. In the personalization setting, the "external corpus" is the past messages and posts of the user, and the retriever's role is to consider representative examples that condition the generator toward the user's voice. Salemi et al. created a benchmark (LaMP) with 7 tasks to test how well AI can personalize responses [2]. Each task gives the model an input (what to respond to) and a user profile (who it is responding for), and the model is expected to use information from the profile to produce a more personalized answer. A model cannot have strong personalization without using personal data, but using personal data raises privacy concerns. A person's writing style is unique enough that it has been demonstrated as a semi-biometric identifier. Word choices (e.g., "big" vs. "enormous"), termed distinctive lexical patterns, the structure of sentences, and idiosyncratic patterns such as punctuation habits, phrases, and tone can be enough to identify a person. Even if an author did not put their name on something, these patterns can be used to figure out who wrote it [3]. This works surprisingly well, even when there are many possible authors to choose from [4]. Recent work has shown that LLMs themselves can be repurposed as highly capable de-anonymization tools, recovering same-author texts across peer reviews, blogs, and emails with alarming precision [5]. If an AI uses an author's real writing style to personalize responses, it carries the risk of stylometric re-identification. To mitigate this, researchers in the field of adversarial stylometry have developed techniques to rewrite text in a neutral style [3]. Making LLMs both highly personal and fully private is inherently conflicting, and this study aims to investigate that conflict precisely.

Despite growing interest in privacy-preserving NLP, prior work has focused predominantly on developing mitigation techniques such as differential privacy, data minimization, and text anonymization as standalone procedures. Comparatively little attention has been paid to directly quantifying the impact of private data suppression on LLM personalization quality — that is, how much personalisation signal is lost when identifying information is removed from the conditioning context. This gap motivates the central inquiry of the present study: to empirically measure the privacy–personalization trade-off within a controlled LLM-based paraphrasing pipeline.

Judging whether a rewritten text still preserves someone's personal style is difficult and not straightforward to measure. Surface-level metrics such as BLEU or ROUGE are poorly suited to judgments about voice, and large-scale human annotation is expensive and slow. Instead of using humans to evaluate text quality, Zheng et al. used another LLM (such as ChatGPT) to act as a judge. The LLM's judgments match human judgments closely [6]; the LLM agrees with humans about as much as humans agree with each other. Typically, human–human agreement is around 80% [6]. We adopt this paradigm here, using two independent judge models from different families (OpenAI's GPT-4.1 and Google's Gemini 2.0 Flash) to mitigate model-specific bias, and complement it with a small human-evaluation study.

The objective of this study is to investigate how personal data affects LLM behaviour when applied for personalization. Specifically, we address the following research questions: (RQ1) How does an LLM respond when it receives private data as a conditioning context? (RQ2) Does the presence or absence of private data measurably impact LLM personalization quality? Together, these questions frame the study as a controlled empirical investigation of the causal relationship between data privacy and personalization effectiveness, using the LaMP-7 Twitter dataset as the experimental testbed.

Our contributions are threefold. First, we show that, with full access to a user's real data, the LLM (Claude Sonnet 4) generated paraphrases achieved near-parity with human-authored ground truth on average (49.7% vs. 50.3%). It is evident that the LLM can produce paraphrases that are nearly indistinguishable from human-written ones. Second, we demonstrate that the proposed PII-suppression procedure retains semantic content at an average rate of 94.8% across judges. Third, the results provide strong evidence of a privacy–personalization trade-off: the features that make a paraphrase sound more like its author are the same features that make the author re-identifiable.

The remainder of this paper is organized as follows. Section 2 reviews related work. Section 3 describes the methodology, including the dataset, profile risk analysis, paraphrasing configurations (P1 and P3), and the profile conversion and verification procedure (P2). Section 4 presents the evaluation results, covering rule-based metrics, LLM-as-a-Judge evaluation for both P1 and P3 with inter-judge agreement analysis, and a human evaluation study, followed by a discussion of the overall findings. Section 5 concludes the paper and outlines directions for future work.

## 2 Related Works

This section surveys four strands of research that converge on the problem addressed in this paper: (i) LLM personalization with retrieval-augmented generation, (ii) privacy risks posed by stylometric authorship attribution, and the countermeasures developed against them, (iii) PII leakage and redaction in large language models, and (iv) evaluation paradigms for personalized text generation, with particular attention to the LLM-as-a-judge framework. We conclude with a synthesis that motivates the specific research gap addressed by the present study.

RAG personalizes LLMs by feeding them relevant user information before they generate a response. Using user profiles significantly improves an LLM's performance, and the LaMP benchmark provides a structured way to measure that improvement [2]. Salemi et al. proposed two optimization algorithms, one based on reinforcement learning with a task-metric reward and another on knowledge distillation from the downstream LLM [7]. Mysore et al. introduced PEARL, a generation-calibrated retriever trained with a scale-calibrating KL-divergence objective that doubles as a performance predictor for long-form personalized writing [8]. In another work, Salemi et al. report that combining RAG with PEFT yields a 15.98% average improvement over non-personalized LLMs on the LaMP benchmark [9].

RAG is particularly beneficial for cold-start users, while PEFT works well when more user-specific data is available [9]. Tan et al.'s One PEFT Per User (OPPU) framework stores user behavior in personalized PEFT modules and outperforms prompt-based baselines across seven diverse LaMP tasks [10].

Writing style can act like a fingerprint, and modern AI systems use detailed linguistic features to identify authors [11]. Deep learning models can now identify who wrote even very short texts (such as tweets) accurately, achieving state-of-the-art performance on short social-media texts across four benchmark datasets [12]. Huertas-Tato et al.'s trained PART model achieves 72.39% zero-shot accuracy on a 250-author task, outperforming RoBERTa embeddings by more than 50 percentage points [13]. Can one still recognize someone's writing style if they write about something totally different? Traditional style features can still identify an author even if they switch topics. Altakrori et al. further show that stylometric features combined with part-of-speech tags are robust to topic shift in ways that large pre-trained language models surprisingly are not [14].

Alongside stylometry research, a different type of research focuses on preventing AI systems from exposing personal identifying information (PII). Lukas et al. formalize three classes of PII leakage via black-box extraction, inference, and reconstruction attacks [15]. Their work shows that naive scrubbing is insufficient and demonstrates that even sentence-level differential privacy still leaks approximately 3% of PII sequences. Kim et al.'s ProPILE tool provides data subjects with a means of probing their own PII exposure in deployed LLM services [16]. Das et al.'s survey shows that modern LLMs can be attacked in multiple ways, for instance by tricking them, corrupting their training data, or making them leak personal information [17]. On the defensive side, Sun et al.'s DePrompt framework uses adversarial generative desensitization to keep the meaning of the text but hide clues that connect it to a specific person or their private data in LLM prompts [18]. In another work, Xiao et al.'s PrivacyMind demonstrates that LLMs themselves can be trained as contextual privacy-protection learners through instruction tuning with positive and negative examples [19].

A major practical difficulty in research on personalized generation is evaluation. Reference-based metrics such as BLEU and ROUGE are poorly aligned with human judgments, and large-scale human annotation is expensive. Zheng et al. introduced the LLM-as-a-judge paradigm and demonstrated that strong LLM judges achieve over 80% agreement with human preferences [6]. However, subsequent work has shown that the approach is not without limitations. Panickssery et al. document a self-preference bias in which LLM judges recognize and favor their own generations [20]. Wataoka et al. measure this effect and show that it is caused by LLMs preferring lower-perplexity text regardless of source [21]; this happens because LLMs tend to give higher scores to more fluent, predictable text, regardless of whether it was written by a human or an AI. Koo et al.'s CoBBLEr benchmark found six different cognitive biases in LLM evaluators [22]. For personalization specifically, Dong et al. find that naive LLM-as-a-Personalized-Judge has low and inconsistent agreement with human ground truth [23]; however, they suggest letting the LLM report its own confidence, and when it is highly confident, its answers are much more reliable (agreeing with others over 80% of the time) [23].

# 3 Method

The methodology of this study is designed to examine the impact of stylometric signal reduction on personalization in large language model (LLM) text generation. A controlled experimental framework (Figure 1) is used with the LaMP-7 benchmark. It integrates profile risk analysis, profile transformation, and multi-stage evaluation. The approach is structured into three phases: (i) generation of personalized paraphrases using original user profiles, (ii) systematic anonymization of profiles to suppress identifiable stylistic and personal features, and (iii) generation of personalized paraphrases using anonymized user profiles. After that evaluation of the resulting outputs using both LLM-based and human judgments. This design enables a direct assessment of the trade-off between privacy preservation and stylistic fidelity in LLM personalization.

## *3.1 Dataset*

In this study, the LaMP-7 (Language Model Personalization) benchmark dataset is used. It contains 1,500 user records taken from the Twitter platform. LaMP-7 is a personalized natural language generation task in which each record represents a user characterized by: (i) a target tweet to be paraphrased, formulated as an explicit paraphrase instruction, and (ii) a personal tweet history. The dataset was originally introduced as part of the LaMP benchmark suite [2] to evaluate the capacity of language models to generate personalized outputs from user-specific context.

A random subsample of 250 records was taken from the 1,500 records using fixed seed 42. Seeding guarantees full reproducibility of all downstream experimental conditions. The 250-record working sample constitutes 16.7% of the full development split and was verified to preserve the distributional characteristics of the parent corpus with respect to profile tweet count.

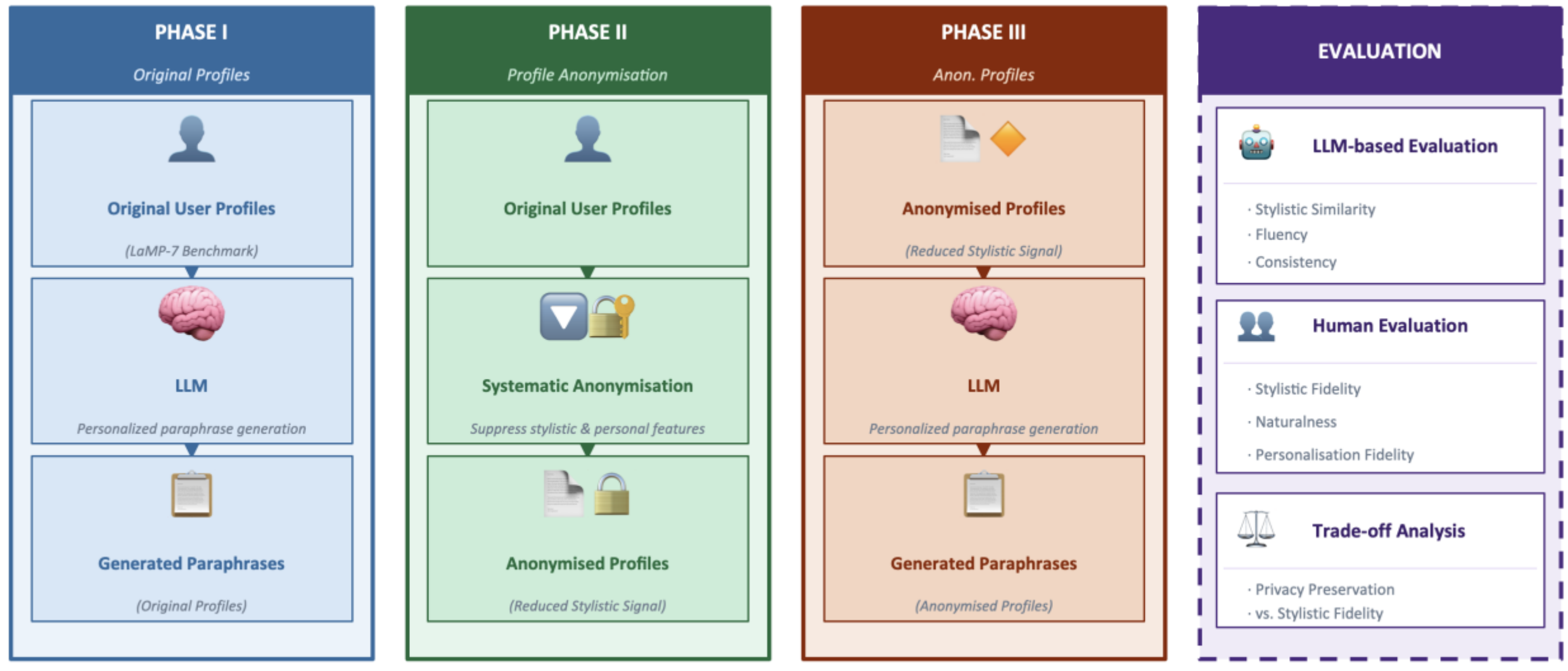

*Figure 1 Methodology overview*

### *3.2 Profile risk analysis*

Prior to anonymization and personalization testing, a profile risk analysis was conducted on the dataset to assess the extent to which conversational data could expose personally identifiable information. To systematically evaluate this risk, five categories of privacy-sensitive signals were identified that are inherently present in tweet conversation data. The first, Age & Demographic Identifiers ($f_1$), encompasses linguistic and contextual cues that may reveal a user's age, gender, or demographic background [24]. The second, Cultural References ($f_2$), captures region- or community-specific allusions embedded in natural language that could narrow a user's cultural or geographic identity [25]. The third, Personal-Life Details ($f_3$), covers any disclosed information relating to an individual's private circumstances, relationships, or lived experiences [26]. The fourth, Niche or Proprietary References ($f_4$), addresses mentions of specialized tools, platforms, or domain-specific knowledge that could uniquely identify a user within a particular professional or interest-based community [27]. Finally, Stylistic Information ($f_5$) encompasses informal language patterns, slang, abbreviations, and filler tokens — surface-level features that have been demonstrated to carry strong authorship signals and pose considerable de-anonymization risks [28, 29]. These five risk factors collectively formed the basis of the profile risk assessment framework applied to the dataset before any further processing.

To quantify the privacy risk embedded in each user profile, a binary flagging scheme was applied across the five risk categories ($f_1$–$f_5$) prior to anonymization. Each conversation profile was assigned a value of 1 if the corresponding risk factor was detected within the user's tweets, and 0 otherwise, resulting in a risk flag vector per profile. The total number of raised flags (#Flags) was then summed across all five dimensions, yielding a profile-level risk score ranging from 0 to 5. To illustrate this process, Sample #610 was examined in detail across two slides. The profile tweets were colour-coded according to the detected risk category — cyan for cultural and media references ($f_2$), red for personal-life details ($f_3$), gold for niche or proprietary references ($f_4$), and blue for informal style and slang ($f_5$). In this sample, no age or demographic identifiers were found ($f_1 = 0$), while all remaining four factors were flagged ($f_2 = f_3 = f_4 = f_5 = 1$), producing a total risk score of 4. This flagging procedure was systematically applied across all profiles in the dataset (as summarized in the flagging in appendix Table 6), where the majority of profiles exhibited between 3 and 5 active flag. The resulting flag violation rates (showed in figure Figure 2) revealed that informal language and slang ($f_5$) was by far the most pervasive risk factor, present in 98.8% of profiles, followed by personal-life details ($f_3$) at 81.6%, niche or proprietary references ($f_4$) at 66.0%, and cultural or pop-culture references ($f_2$) at 62.0%. Age and demographic identifiers ($f_1$) were the least frequently occurring signal, flagged in only 18.8% of profiles. These findings confirm that raw tweet conversation data carries a substantial and multi-dimensional privacy risk, with stylistic and personal-life signals being particularly ubiquitous, underscoring the necessity of a comprehensive anonymization strategy prior to any downstream analysis.

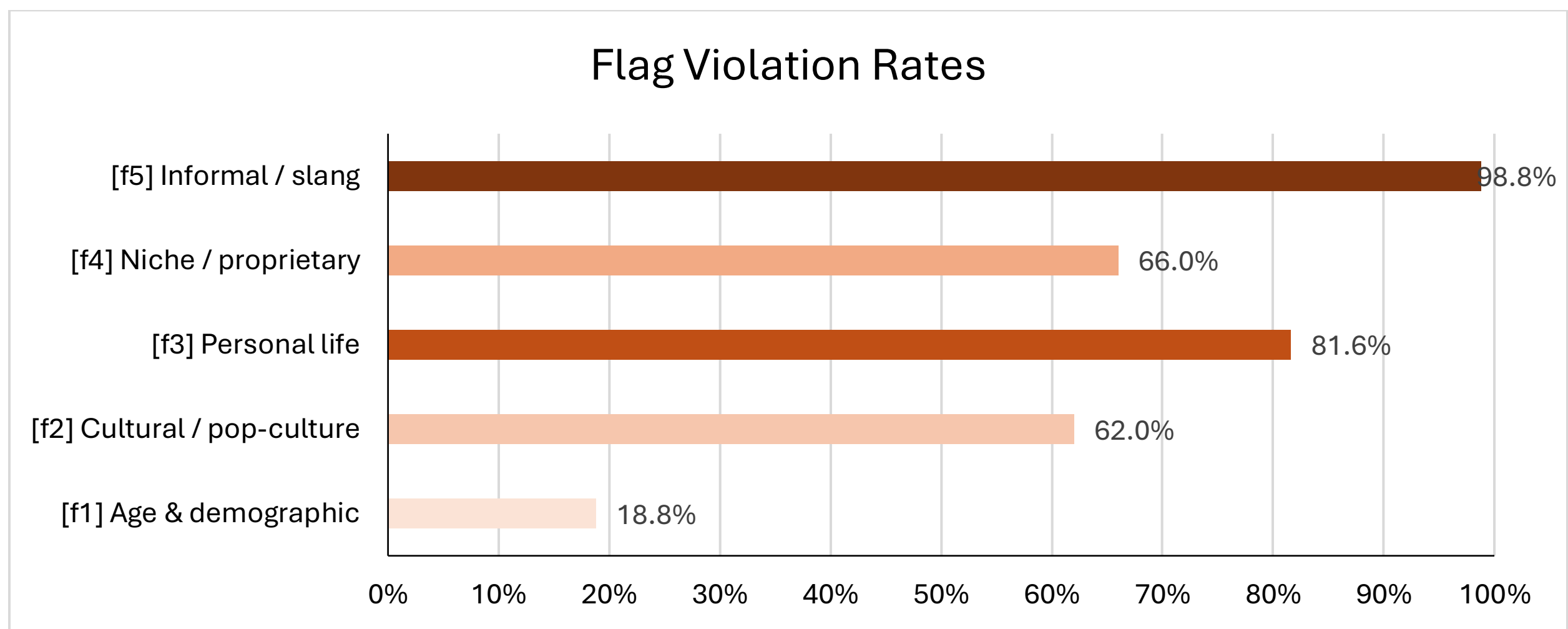

*Figure 2 Flag violation rates across five privacy risk categories ($f_1$–$f_5$) in the tweet conversation dataset prior to anonymization.*

To move beyond a simple flag count and produce a more nuanced measure of privacy exposure, a weighted privacy-risk score S was computed for each profile. Formally, letting $f_i \in \{0, 1\}$ denote the binary flag for risk category i and $w_i$ the corresponding weight, the score is defined as:

$$S = \frac{\sum_{i=1}^{5} w_i \cdot f_i}{\sum_{i=1}^{5} w_i}$$

where $S \in [0, 1]$, with $S = 0$ indicating full compliance and $S = 1$ indicating that all risk rules are violated. The weights $\{w_1, w_2, w_3, w_4, w_5\} = \{1, 2, 4, 3, 5\}$ were assigned to reflect the relative severity of each risk category, with stylistic fingerprints ($f_5$) receiving the highest weight of 5 — accounting for 33.3% of the total weighted score — given their well-documented capacity for de-anonymization. The age and demographic identifiers ($f_1$) received the lowest weight of 1 (6.7%), as they are less consistently present in conversational tweet data. Personal-life details ($f_3$) were assigned a weight of 4 (26.7%), niche and proprietary references ($f_4$) a weight of 3 (20.0%), and cultural and media references ($f_2$) a weight of 2 (13.3%), with the total weight summing to 15. Applying this scoring formula across the dataset produced risk level classifications into three tiers: HIGH ($S \geq 0.60$), MEDIUM ($0.30 \leq S < 0.60$), and LOW ($S < 0.30$). The results revealed a strikingly skewed distribution (showed in figure Figure 3), with 87% of profiles classified as HIGH risk, 12% as MEDIUM, and only 1% as LOW. This outcome strongly demonstrates that the vast majority of raw tweet conversation profiles carry a substantial privacy risk across multiple weighted dimensions.

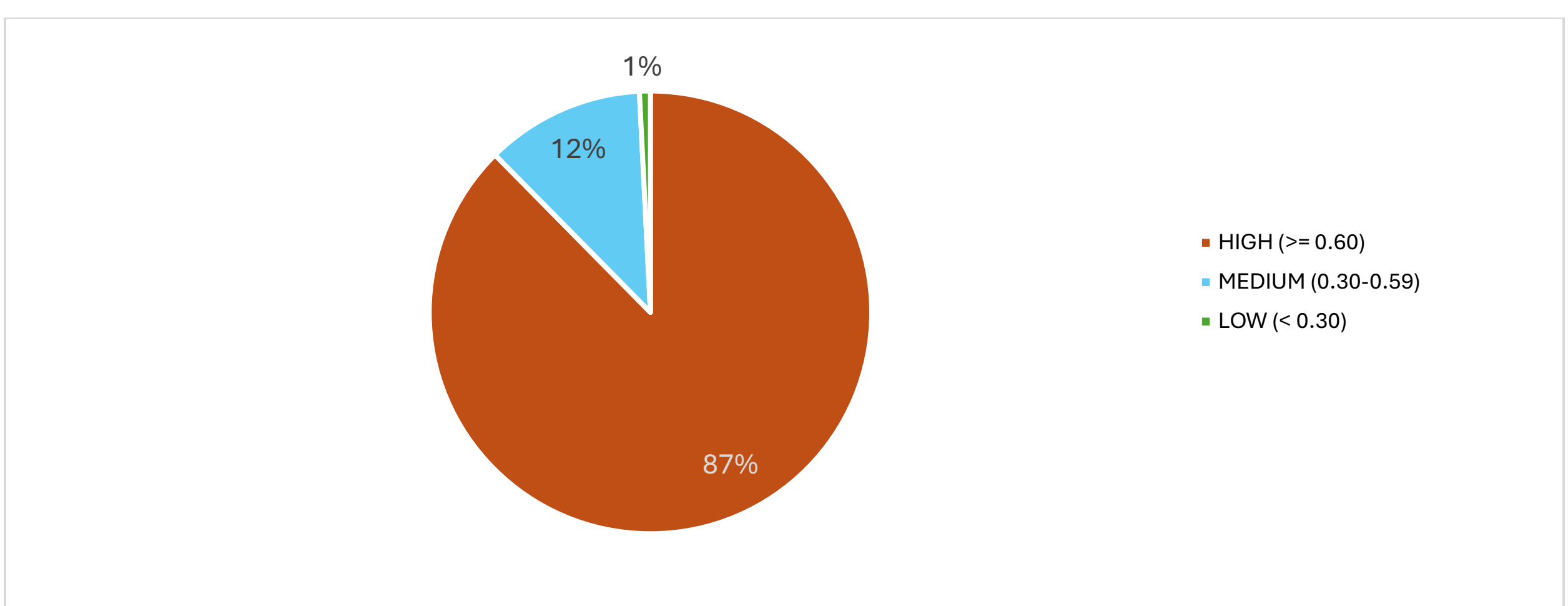

*Figure 3 Risk level distribution in the dataset*

### *3.3 Paraphrase Tweet Based on Original Profile (P1)*

Following the privacy risk analysis stage, the first personalization configuration (P1) was applied, in which the original tweet profile is used as the reference context to paraphrase each candidate tweet. Two models of contrasting capability tiers were employed for this task: Claude (Anthropic, claude-sonnet-4), a high-tier large language model, and Google's FLAN-T5 Base, a lightweight low-tier model. Both models were tasked with generating stylistically modified versions of the candidate tweet conditioned on the user's profile context. The rationale behind employing two models of deliberately different scales was to examine and compare how model capacity influences personalization behaviour. Specifically, whether a high-tier model produces richer, more contextually aligned paraphrases compared to a low-tier model.

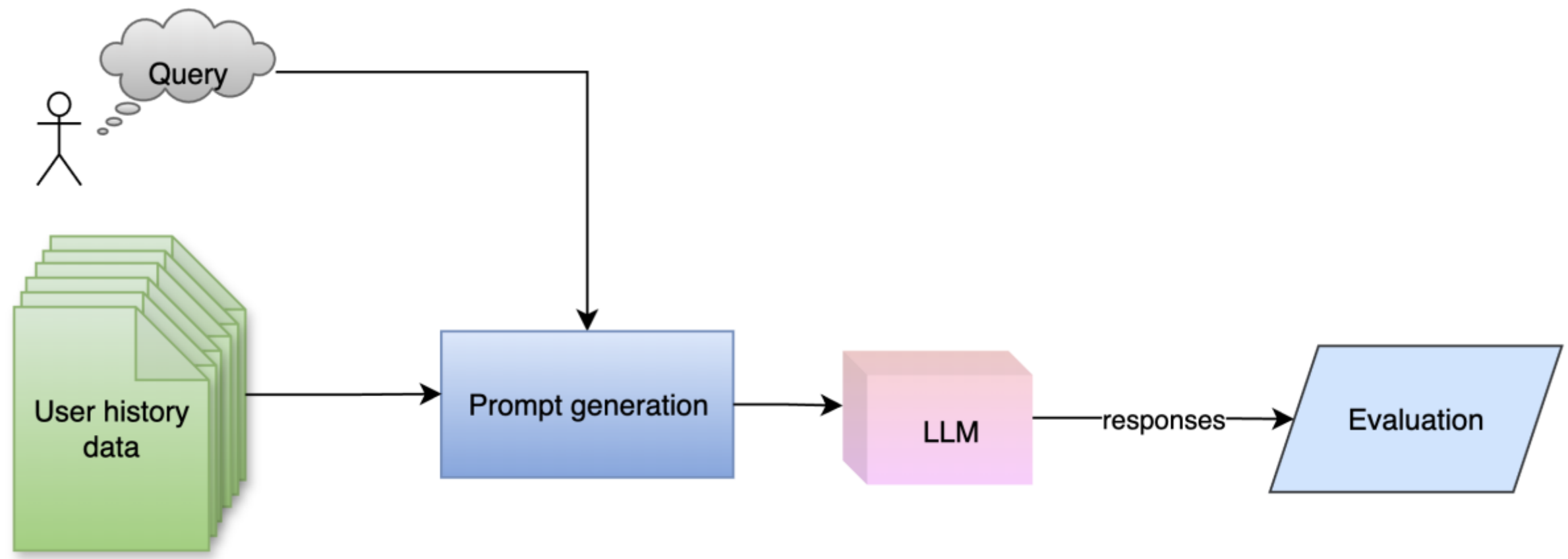

*Figure 4 Paraphrase process with original tweet profile.*

The generator is used strictly in inference mode. Each evaluation row pairs a single source tweet with its associated profile string. Both are loaded from a tabular input file and combined into a minimal instruction-following prompt that constrains the model to return only the paraphrased tweet. Generation is performed with Claude Sonnet 4 through the Anthropic Python SDK, and Google FLAN-T5 Base through the HuggingFace Transformers library. Both models' default decoding parameters were used so that the results reflect their out-of-the-box behaviour rather than any task-specific tuning. The use of two models of deliberately different capability tiers a high-tier large language model and a lightweight low-tier model was for a systematic comparison of how model capacity influences personalization behaviour.

*Paraphrase the following tweet without any explanation before or after it according to the profile:*

*{tweet}*

*Profile:{profile}*

Three features of this prompt are intentional. First, the instruction is phrased negatively with respect to auxiliary output ("without any explanation before or after it"), which constrains the generator to emit only the paraphrase and removes the need for any post-processing or parsing step. Second, the word *Profile* is reused verbatim as a block header so that the generator is cued to treat the subsequent text as a reference document describing the user rather than as further instruction. Third, the tweet is placed before the profile, aligning the most salient span of the prompt with the position of the instruction's direct object. No few-shot exemplars, retrieved context, or explicit style descriptors are injected: the profile string itself, produced upstream by the retrieval and conversion stages, carries the full personalisation signal.

### *3.4 Profile Conversion and Context Verification (P2)*

Profile paraphrasing was performed using Claude (Anthropic, claude-sonnet-4) as the generation model, applied to the LaMP-7 Twitter dataset. Each user's combined tweet history was passed to the model, with a structured system prompt encoding five anonymization constraints: (i) removal of age and demographic identifiers; (ii) generalization of specific cultural references (e.g., celebrity names, television programmes) into neutral terms; (iii) redaction of personal life details, including family circumstances and location-specific content; (iv) substitution of niche or proprietary references with broad equivalents; and (v) conversion of informal language, slang, and abbreviations into formal prose. The model was instructed to return only the paraphrased output, without preamble or explanation. Prompt is provided in the appendix Table 13.

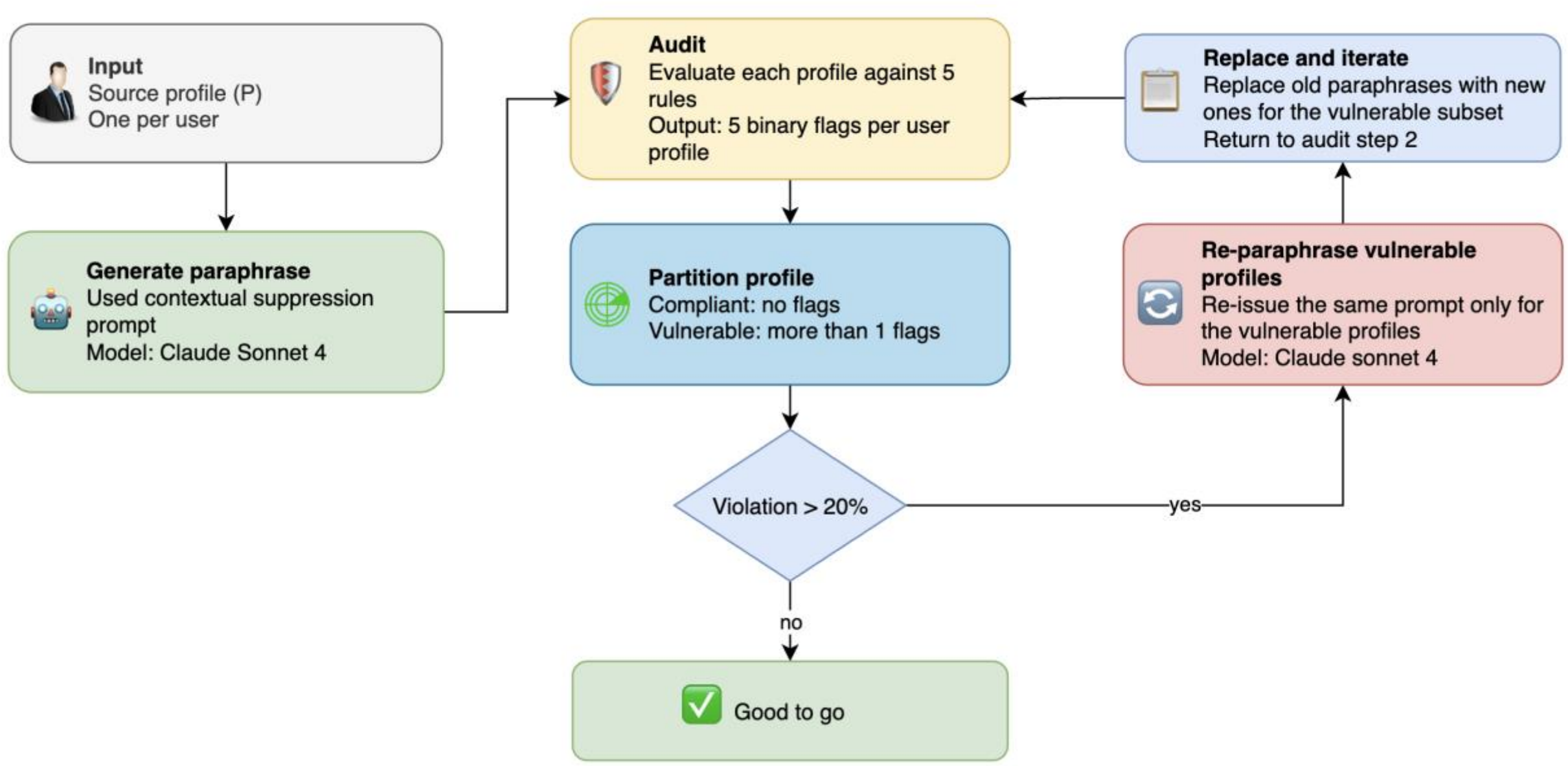

*Figure 5 Iterative refinement loop for privacy-preserving profile paraphrasing*

To verify the completeness of the suppression step, each converted conversation was subsequently audited by an independent model — Gemini 2.5 Flash — acting as an adversarial reviewer. The reviewer was prompted to identify any residual personally identifiable or stylistically distinctive content that remained after the initial paraphrasing pass. Conversations flagged by the reviewer as still carrying such content were re-submitted to the original Claude paraphrasing prompt for a second pass.

The iterative procedure is defined over the following notation. Let:

- ***u*** — a user with combined tweet history $C_u^{(0)}$ (the original profile);

- $G$ — the generator model (Claude claude-sonnet-4);
- $A$ — the auditor model (Gemini gemini-2.5-flash);
- $\pi_{conv}$ — the conversion (paraphrasing) prompt with the five anonymisation constraints;
- $\pi_{audit}$ — the audit prompt;
- $F_t$ — the set of residual PII or stylistic flags returned by the auditor at iteration t;
- $T_{max}$ — the maximum number of revision iterations permitted;
- $C_u^*$ — the final converted profile for user u used in the P3 evaluation.

The conversion procedure for each user u is defined by the iteration

$$C_u^{(t+1)} = G(\ \pi_{conv} \ \|\ C_u^{(t)} \ \|\ F_t\ ), \qquad F_t = A(\ \pi_{audit} \ \|\ C_u^{(t)}\ )$$

with stopping condition

$$C_u^* = C_u^{(t^*)}, \qquad t^* = \min\{\ t \in \{0, 1, \ldots, T_{max}\} :\ F_t = \emptyset\ \}$$

where || denotes prompt concatenation and $F_t = \emptyset$ indicates that the auditor returned no further flags.

This two-stage procedure was adopted to reduce the risk that residual identifying signal would persist in the converted profiles due to blind spots specific to the generator. Another objective of this is to ensure the maximum amount of personally identifiable information had been removed before the converted profiles were used in the P3 evaluation. Psudocode is provided in appendix Table 14.

The two-pass iterative procedure yielded a substantial reduction in privacy-sensitive signals across all five risk categories, as illustrated in Figure 6. Flag violation rates dropped markedly following anonymization: cultural and pop-culture references ($f_2$) fell from 62.0% to 2.8%, niche and proprietary references ($f_4$) from 66.0% to 3.2%, and informal language and slang ($f_5$) — the most pervasive signal prior to anonymization at 98.8% — was fully suppressed to 0.0%. Age and demographic identifiers ($f_1$) also declined from 18.8% to 11.6%. Personal-life details ($f_3$) exhibited the greatest resistance to neutralization, reducing from 81.6% to 48.4%, suggesting that deeply embedded personal narratives present a persistent challenge even under iterative refinement. Some personal information still remains because people express personal context indirectly in natural conversation. Completely removing those clues is difficult without damaging the meaning and flow of the text. Even so, the two-step pipeline worked well overall. Four of the five privacy-risk categories were almost completely removed, showing that the converted profiles were anonymous enough to be safely used in the later P3 personalization experiments.

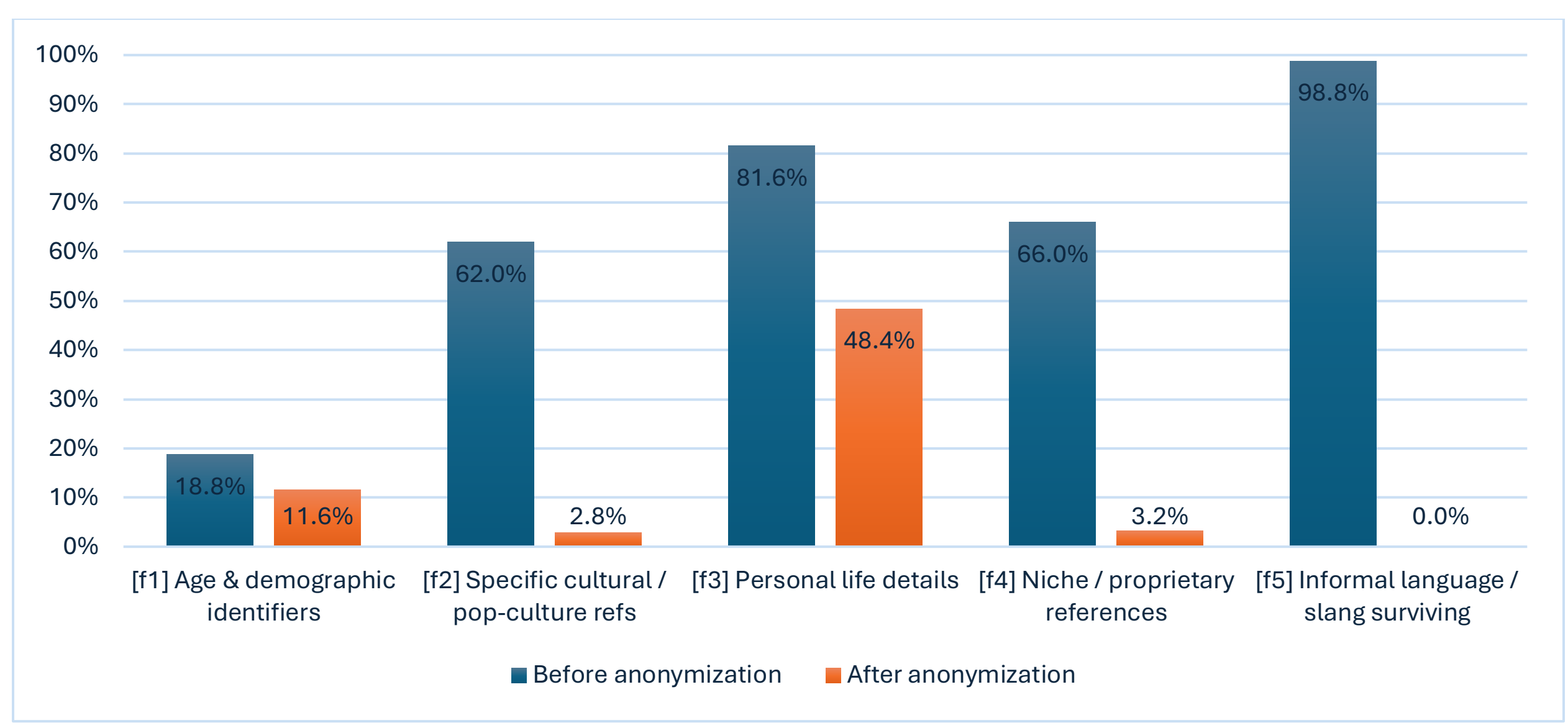

*Figure 6 Flag violation rate (Before vs After Anonymization)*

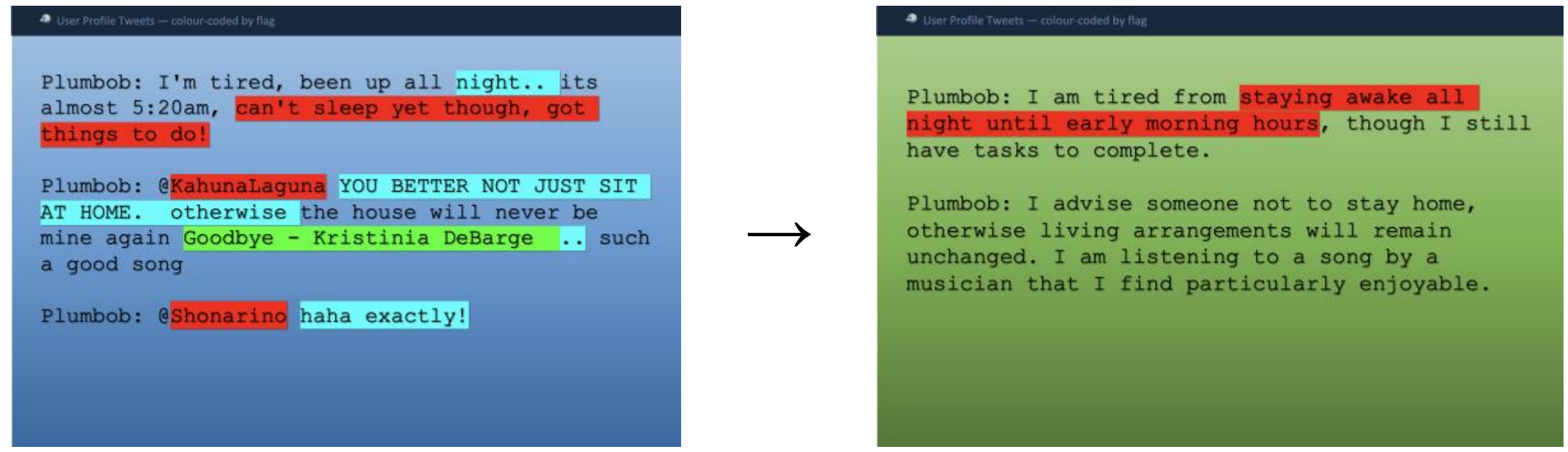

*Figure 7 Before and after anonymization sample*

### *3.4.1 Context Verfication*

The quality of the resulting paraphrases was assessed in Phase 2 through evaluation tasks. The task assessed semantic context preservation, requiring judges to produce a binary label (Intact: Yes/No) indicating whether the Claude paraphrase retained the contextual meaning of the original profile. The aggregated results are reported in Table 1. Following is the prompt provided to the LLMs for evaluation.

> *You are an expert at analyzing conversation context and semantic fidelity.You always respond in this exact format, nothing else:*
>
> *INTACT: <Yes or No>*
>
> *ORIGINAL_CONTEXT: <one sentence describing the core context of the original>*
>
> *PARAPHRASED_CONTEXT: <one sentence describing the core context of the paraphrased version>*

*Table 1 Semantic Context Preservation Results.*

| Judge Model | Intact (Yes) | Intact Rate | Not Intact (No) | Failure Rate | Total |
|---|---|---|---|---|---|
| OpenAI GPT-4.1 | 244 | 97.6% | 6 | 2.4% | 250 |
| Gemini 2.0 Flash | 230 | 92.0% | 20 | 8.0% | 250 |
| Average | 237 | 94.8% | 13 | 5.2% | 250 |

*Intact (Yes) indicates the Claude-generated paraphrase successfully preserved the contextual meaning of the original tweet profile.*

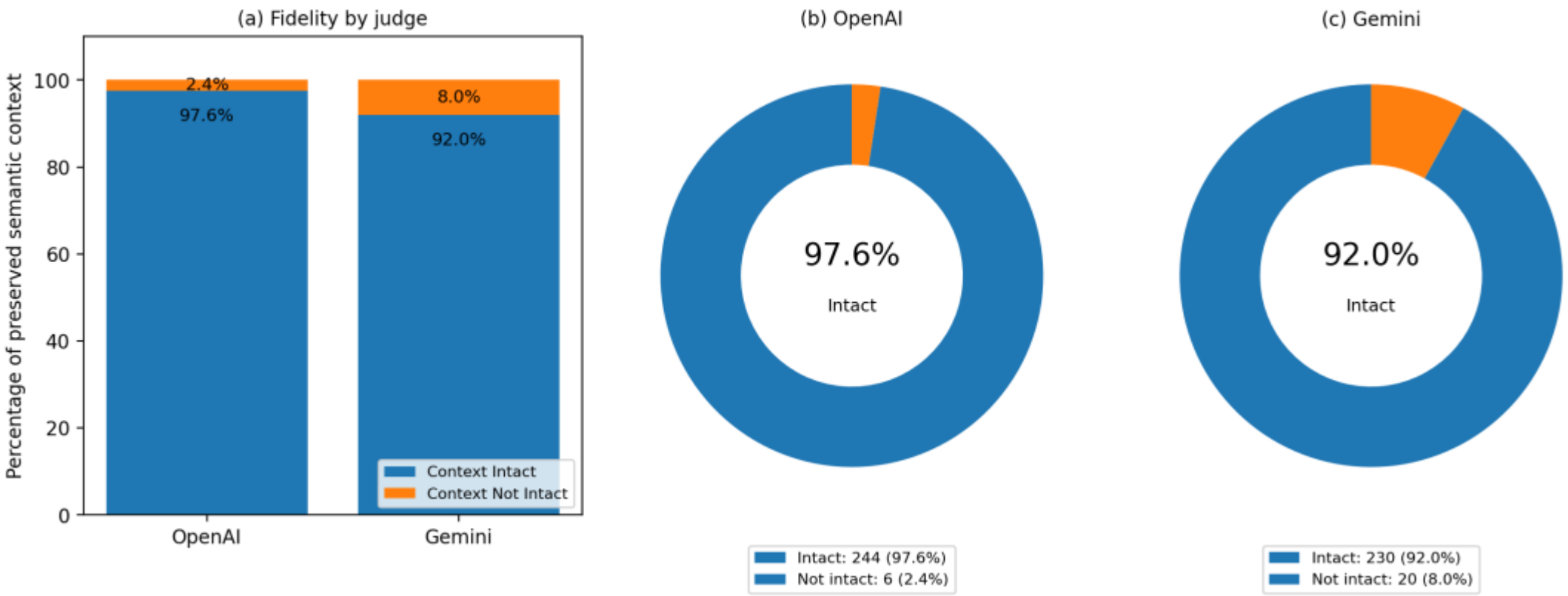

*Figure 8 Context fidelity of paraphrases.*

Context preservation results were consistently high across both judges. OpenAI GPT-4.1 labelled 244 of 250 paraphrases as semantically intact, leaving only six failure cases (2.4%). Gemini 2.0 Flash applied a slightly stricter criterion, marking 230 of 250 paraphrases (92.0%) as intact and flagging 20 (8.0%) as failures. Averaged across the two judges, 94.8% of the generated paraphrases were judged to preserve the semantic content of the source profile, with a corresponding average failure rate of 5.2%. The 5.6 percentage-point gap between the two judges is consistent with the broader pattern observed throughout this evaluation, in which Gemini 2.0 Flash adopts more conservative judgements than GPT-4.1. The gap is substantially smaller in this binary fidelity task than in the pairwise personalisation comparison reported earlier. This observation suggests that the judges agree more closely on whether meaning has been preserved.

Figure 8 visualises these results, showing the per-judge distribution of intact and not-intact labels alongside the cross-judge average and making the consistency of the fidelity finding immediately apparent.

### *3.5 Paraphrase Tweet Based on Converted Profile (P3)*

Following the profile conversion stage described in Section 3.4, we applied the P3 configuration. In this phase the converted (PII-suppressed) tweet profile is used as the reference context to paraphrase each candidate tweet. The paraphrasing was performed using Claude (Anthropic, claude-sonnet-4) and Google FlanT5, which generated stylistically modified versions of the candidate tweet conditioned on the anonymised profile rather than the original.

The intent of this stage is to establish a privacy-preserving counterpart to P1. Whereas P1 conditions the paraphrase generator on the user's original profile, P3 substitutes the converted profile produced by the PII-suppression procedure of Section 3.4, holding every other component of the pipeline fixed so that the difference between the two configurations isolates the effect of stylistic anonymisation.

To keep inference cost tractable and to align with the retrieval and P1 experiments, the pipeline is restricted to the same 250 rows of the input file used in P1, ensuring matched-pair comparability between the two conditions. The prompt is deliberately minimal and structurally identical to the P1 prompt, differing only in the contents of the profile field:

*Paraphrase the following tweet without any explanation before or after it according to the profile:*

*{tweet}*

*Profile: {converted_profile}*

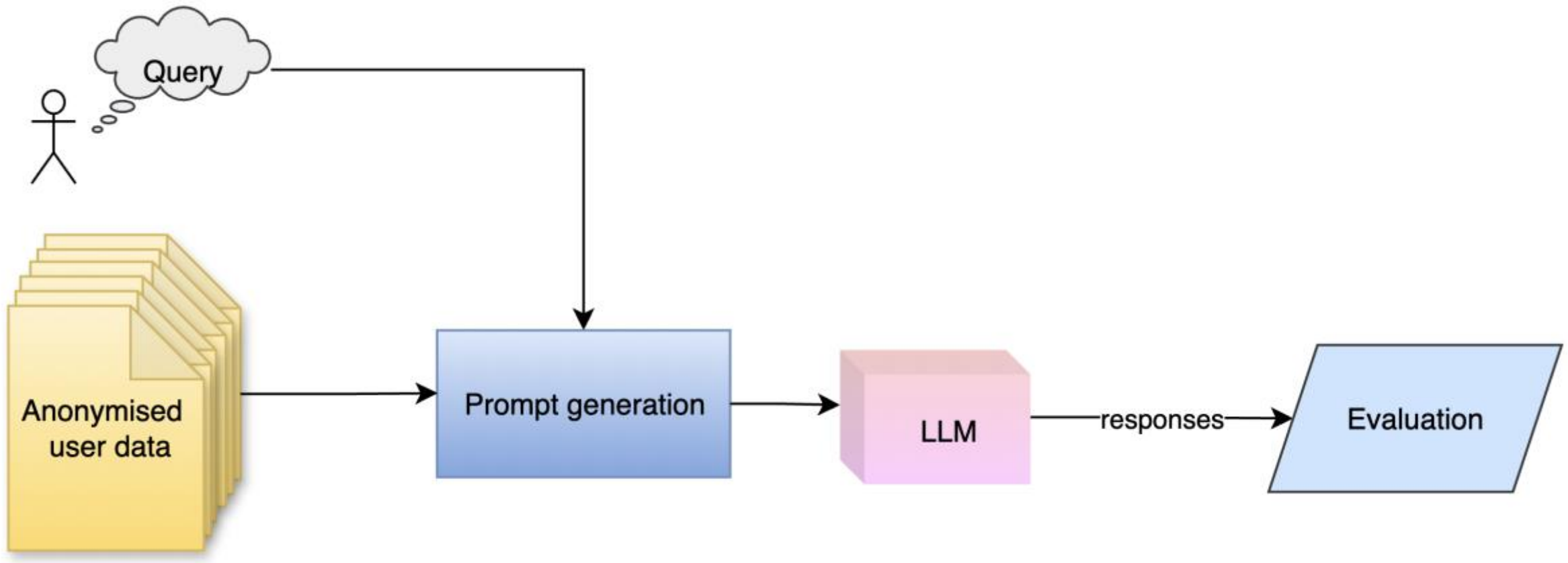

*Figure 9 Paraphrase process with anonymized tweets.*

# 4 Evaluation Analysis

This section presents the empirical evaluation of the personalisation pipeline along two complementary axes. Section 4.2 use two established rule based metrics to evaluate the performance of LLMs based on the ground truth. Sections 4.3 and 4.4 use a dual-judge LLM-as-a-judge protocol (OpenAI GPT-4.1 and Gemini 2.0 Flash) to compare paraphrases generated under the original-profile condition (P1) and the anonymized conversation condition (P3) against their human-authored ground truth. Both process have the aggregate level and through inter-judge agreement analysis. Section 4.5 complements these automated results with a small human-evaluation study that provides an independent, model-free check on the same comparison. Together, these analyses isolate and quantify the personalisation cost of stylistic anonymisation.

## *4.1 Evaluation metric*

***Rule based evaluation metric***

Alongside the LLM-as-a-judge evaluation, two established reference-based metrics (ROUGE [30] and METEOR [31]) were used to quantitatively measure the lexical and semantic similarity between the generated paraphrases and the human-authored ground truth. ROUGE-1 measures unigram overlap between the generated output and the reference text, capturing surface-level lexical similarity through precision, recall, and F1 score. ROUGE-L extends this by measuring the longest common subsequence between the two texts. It used to capturing sentence-level structural similarity and word order without requiring consecutive matches. Unlike ROUGE, which mainly counts overlapping words, METEOR rewards synonyms, morphological variants, meaning-preserving word substitutions.

***LLM-as-a-Judge Evaluation***

To assess the quality and stylistic fidelity of the paraphrased outputs, we adopted the LLM-as-a-judge paradigm, in which a large language model is prompted to compare two candidate texts and select the one that better satisfies a given criterion. Zheng et al. showed that capable LLM judges can reach agreement rates exceeding 80% with human annotators, a level comparable to the agreement observed among human experts themselves, which supports the use of LLMs as a practical basis for automated evaluation [6].

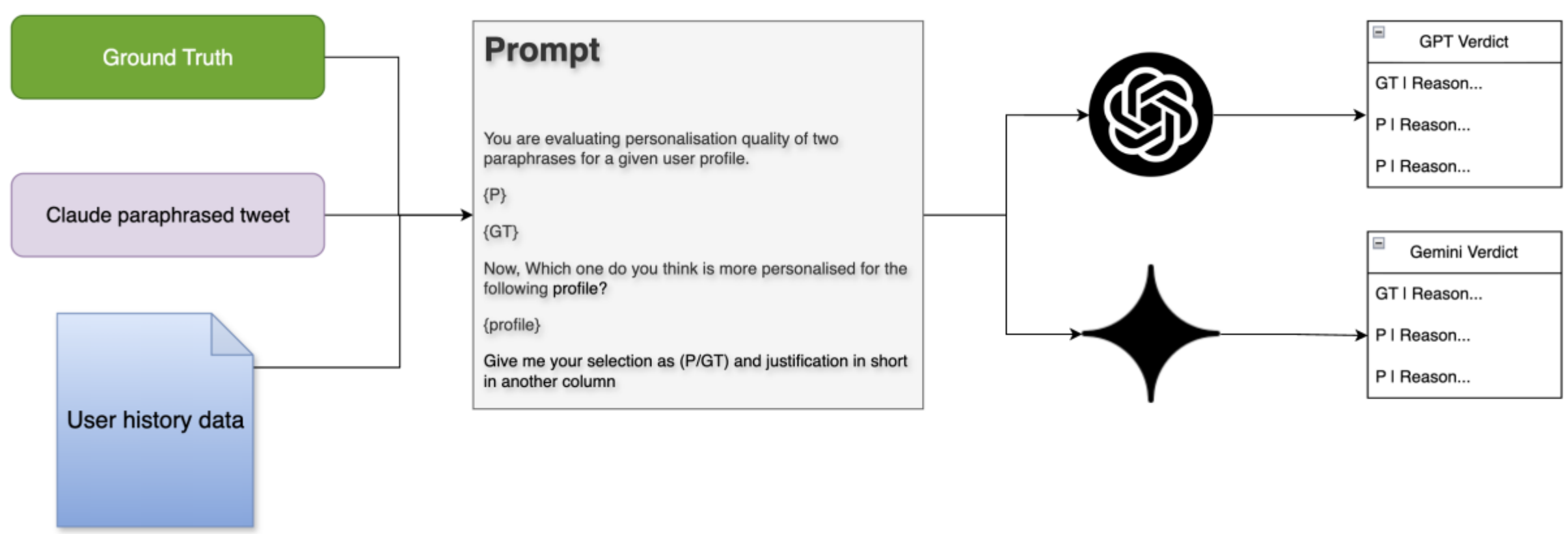

*Figure 10 LaaJ evaluation procedure*

Using multiple judges makes results more reliable: if both models agree, the evaluation is more trustworthy. Building on this finding, we used two independent judge models, OpenAI GPT-4.1 and Google Gemini 2.0 Flash. Employing two judges from different model families helps reduce the risk of model-specific bias and provides a more balanced view of the evaluation outcomes.

## 4.2 Rule based evaluation

### 4.2.1 Rule based evaluation for P1

Figure 11 and Figure 12 illustrate the distribution of ROUGE-1, ROUGE-L, and METEOR scores for Claude and FlanT5, respectively, evaluated on P1 configuration.

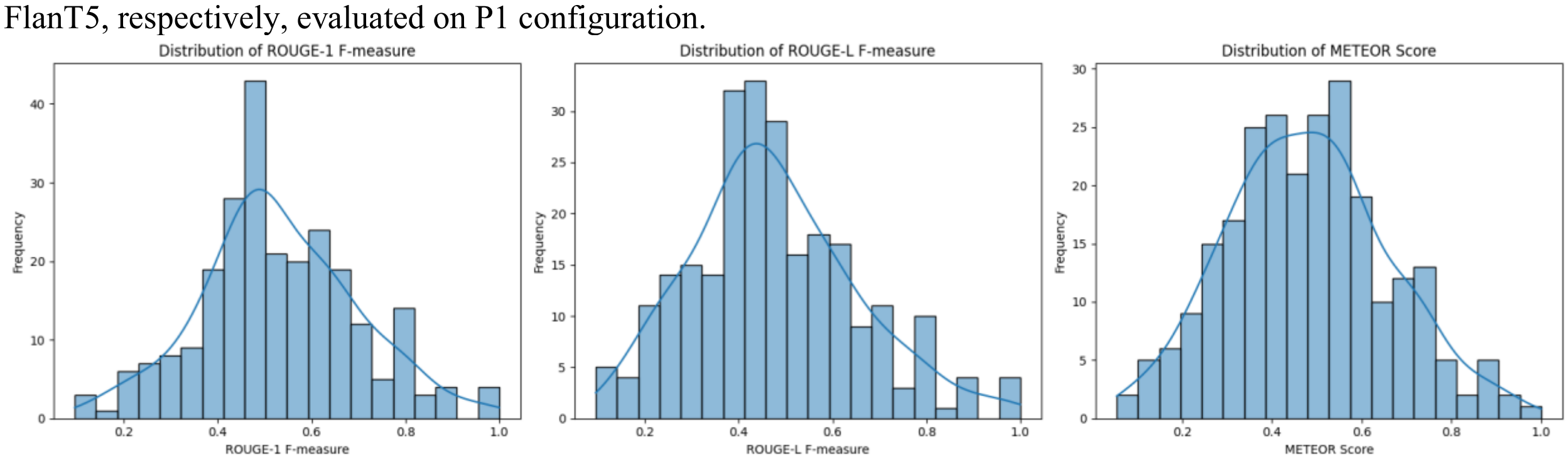

*Figure 11 Distribution of Claude performance for P1*

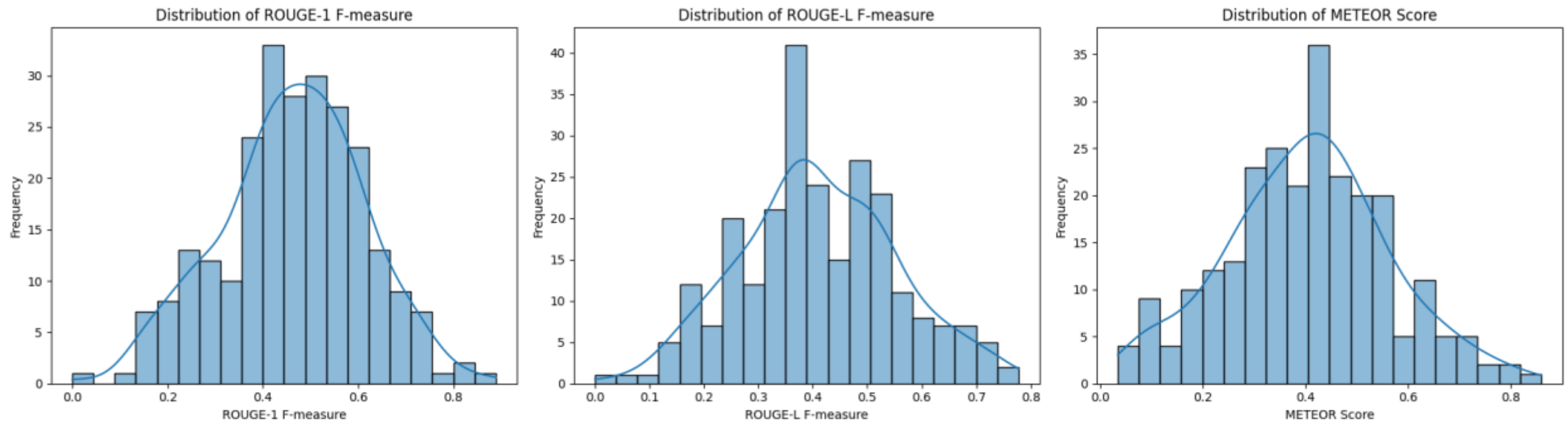

*Figure 12 Distribution of FlanT5 performance for P1*

As shown in Figure 11, Claude's performance distributions across all three metrics exhibit notable variability and a right-skewed pattern, with scores spreading broadly across the 0.2–1.0 range. This shows that while Claude occasionally produces outputs with high lexical overlap with the ground truth, its performance is inconsistent, reflecting sensitivity to input variation or paraphrase diversity in its generated responses. In contrast, Figure 12 reveals that FlanT5's score distributions are more centralized and tightly concentrated, particularly in the 0.3–0.5 range across all three metrics. The narrower spread and more symmetric bell-shaped curves indicate greater consistency in output generation. However, the lower score range also suggests that FlanT5's outputs tend to have less lexical and semantic overlap with the ground truth. This idicates that its paraphrases deviate more uniformly from the reference text rather than occasionally achieving high similarity as seen with Claude.

#### *4.2.2 Rule based evaluation for P3*

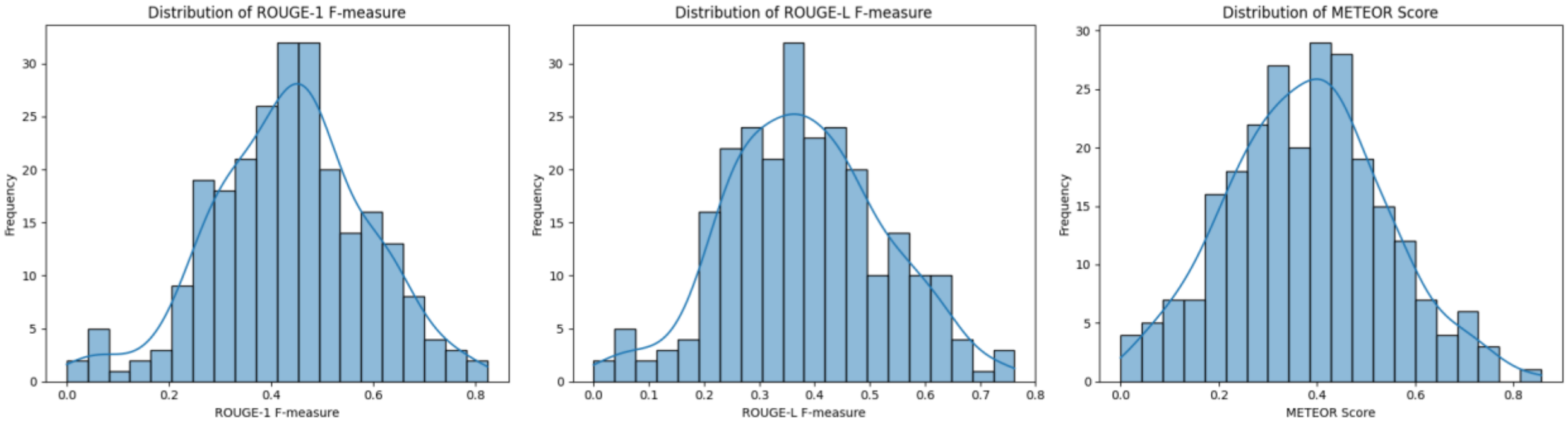

*Figure 13 Distribution of Claude performance for P3*

Under P1 (Figure 11), the distributions across all three metrics exhibit a relatively wide spread with a right-skewed pattern. This spread is indicating considerable variability in the degree of lexical overlap between Claude's generated paraphrases and the ground truth. The presence of scores extending toward higher values (up to 1.0) suggests that some paraphrases achieved strong surface-level similarity with the reference text.

In contrast, the P3 distributions (Figure 13) display a noticeable shift toward the center, with score concentrations tightening around the 0.3–0.5 range across all metrics. This centralization reflects a reduction in output variance following the anonymization process. Importantly, the narrowing of the score distributions indicates that overlap between the generated paraphrases and the ground truth has decreased post-anonymization. It is an expected outcome as anonymization modifies or removes personally identifiable and contextually specific expressions that may have previously contributed to higher similarity scores.

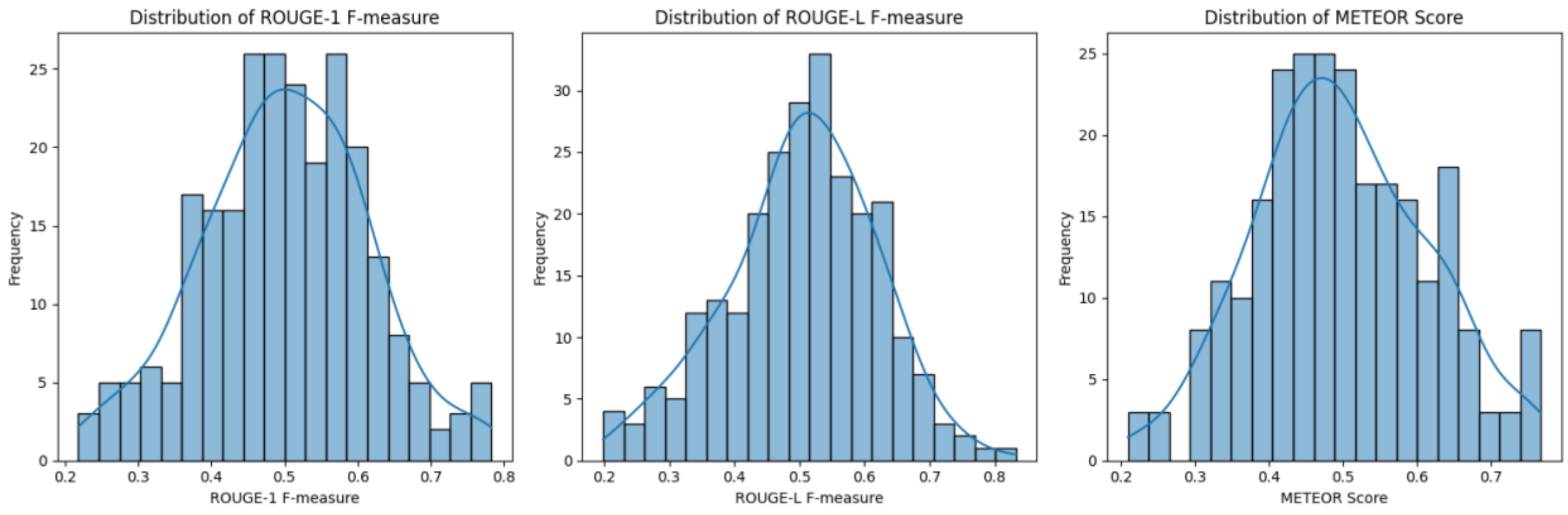

*Figure 14 Distribution of FlanT5 performance for P3*

However, for FlanT5 evaluation, it showed that, it performed average and there is not big difference between P1 and P3 evaluation. It indicates that, FlanT5 did not capture much of the private data to generate the responses.

### *4.3 LaaJ Evaluation of P1 (Paraphrase tweet based on original profile)*

A pairwise evaluation framework was applied across 250 samples from the LaMP-7 dataset to assess the quality of the Claude-generated and FlanT5 generated tweet paraphrases. Two independent LLM judges—OpenAI 4.1 and Gemini 2.0 Flash—were used to ensure that the assessment was not subject to the potential biases of a single model. For each sample, both judges were presented with three inputs: (i) the Claude-generated paraphrase based on the converted profile, (ii) the human-authored ground truth paraphrase, and (iii) the user's original unmasked tweet profile. Each judge was then prompted to identify which of the two paraphrases was more personalised for the given user, responding with an explicit selection—either (Para) for the model-generated output or (GT) for the ground truth—accompanied by a brief justification. Specifically, the judges were prompted as follows:

*"You are evaluating personalization quality of two paraphrases for a given user profile. Which one is more personalized for the following profile?*
*Respond with exactly: (Para) or (GT), followed by a short justification."*

The pairwise evaluation results for P1 are summarised in Table 2 and visualised in Figure 15. Gemini 2.0 Flash selected the Claude-generated paraphrase in 106 of 249 cases (42.57%) and the ground-truth tweet in 143 cases (57.43%), exhibiting a moderate preference for the human-authored output. OpenAI 4.1 produced the inverse pattern, selecting the Claude-generated paraphrase in 142 of 250 cases (56.80%) and the ground truth in 108 cases (43.20%). Averaged across both judges, the ground truth was preferred in 50.31% of cases and the Claude-generated paraphrase in 49.69% a difference of fewer than one selection out of 250 — indicating that at the aggregate level the two sources are statistically indistinguishable.

*Table 2 Pairwise evaluation results for P1.*

| Judge | GT Selected | GT Rate | Model Selected | Model Rate | Total |
|---|---|---|---|---|---|
| **P1 – (Claude Sonnet 4 Paraphrase)** | | | | | |
| OpenAI GPT-4.1 | 108 | 43.20% | 142 | 56.80% | 250 |
| Gemini 2.0 Flash | 143 | 57.43% | 106 | 42.57% | 249 |
| Average | 125.5 | 50.31% | 124 | 49.69% | 249.5 |
| **P1 – (FlanT5 Paraphrase)** | | | | | |
| OpenAI 4.1 | 155 | 62.25% | 94 | 37.75% | 249 |
| Gemini 2.0 Flash | 226 | 90.76% | 23 | 9.27% | 249 |
| Average | 190.0 | 76.46% | 58.5 | 23.54% | 248.5 |

*GT = human-authored ground truth; Model = Claude-generated paraphrase. Each judge evaluated 250 samples from the LaMP-7 dataset.*

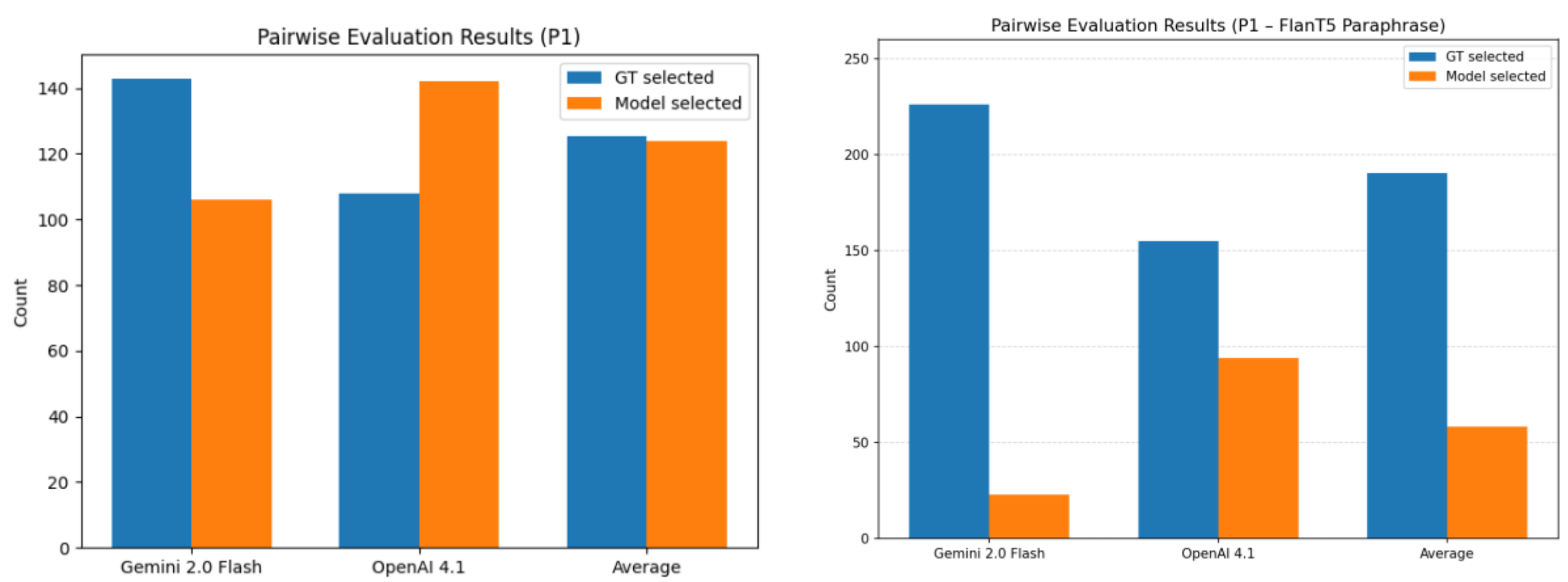

*Figure 15 Pairwise evaluation results for P1: (a) Claude Sonnet 4, (b) Google FlanT5.*

For the FlanT5 generation (Table 2), Gemini 2.0 Flash selected the ground-truth tweet in 226 of 249 cases (90.8%), exhibiting a pronounced and consistent bias toward human-authored output. OpenAI 4.1, by contrast, selected the ground truth in only 155 of 249 cases (62.2%), demonstrating a far more balanced, though still GT-leaning, judgement pattern. The asymmetry in disagreement is particularly revealing: of the 99 cases where the two judges disagreed, 85 (86%) followed the pattern of Gemini selecting the ground truth while OpenAI selected the model output. This directional skew indicates that the disagreements are not random noise but rather reflect a systematic difference in how the two judges perceive the quality of Flan-T5 paraphrases.

#### *4.3.1 Inter-Judge Agreement Analysis*

While the aggregate per-judge rates already point to a near-parity result, they do not reveal how the two judges arrived at this aggregate, nor whether the cases on which the model was preferred were the same cases across judges. To address this, we analysed inter-judge agreement on the 249 samples for which both judges produced valid preferences. The results are presented Figure 16.

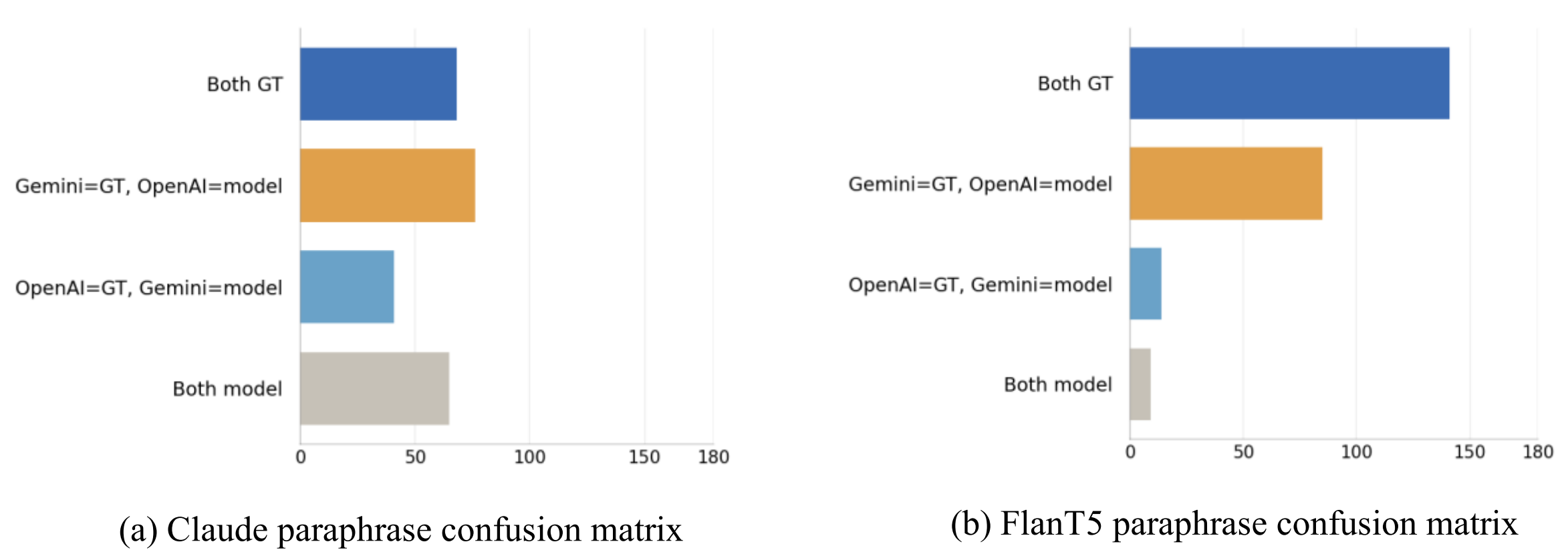

(a) Claude paraphrase confusion matrix

(b) FlanT5 paraphrase confusion matrix

*Figure 16 Inter-judge agreement confusion matrices comparing model paraphrases against ground-truth (GT) selections.*

For Claude paraphrases (in Figure 16), disagreements are spread across all four cells, neither judge dominates, and the two genuinely diverge. For FlanT5, almost all mass concentrates in ”Both GT” (141 items) and ”Gemini=GT, OpenAI=model” (85 items).

The per-item agreement heatmap (Figure 17) shows one small square per item (the first 200 items, displayed side by side for each generator). Each square’s color encodes the combination of choices made by the two judges for that exact item. Dark blue — both judges selected the ground truth (GT), Orange — Gemini selected GT while OpenAI selected the model/paraphrase, Light blue — OpenAI selected GT while Gemini selected the model/paraphrase, Gray — both judges selected the model/paraphrase.

Rather than presenting only a summary statistic, the visualization exposes the raw pattern of agreement and disagreement on an item-by-item basis. The FlanT5 heatmap is dominated by dark blue squares, indicating that both judges defaulted to GT for nearly every item, with only occasional orange regions where Gemini preferred GT but OpenAI disagreed. In contrast, the Claude heatmap appears substantially more heterogeneous, with all four colors occurring frequently. This visual diversitysuggests that the judges were genuinely divided on many individual items rather than merely differing in aggregate statistics.

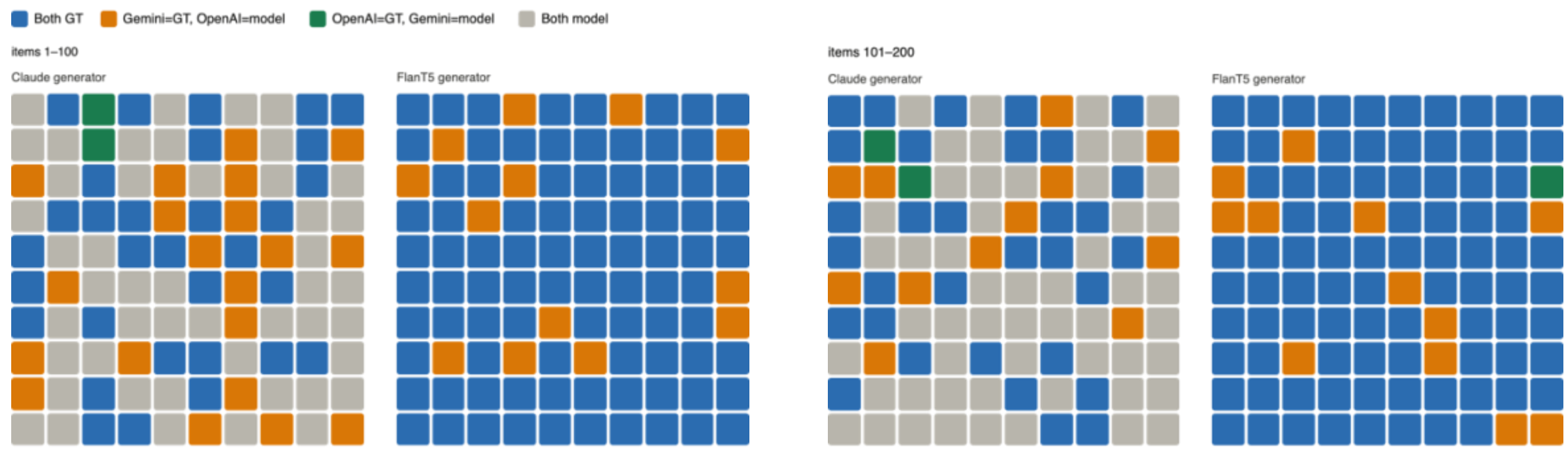

*Figure 17 Per-item agreement heatmap for P1*

In summary, the P1 evaluation establishes that, when the generator is conditioned on the user's unmodified profile, the Claude-generated paraphrase and the human-authored reference are statistically indistinguishable at the aggregate level (49.69% vs. 50.31%), and remain so within the 53.0% subset on which the two judges agree (26.1% Model vs. 26.9% GT). The asymmetric disagreement pattern between the two judges is consistent with documented self-preference and surface-fluency biases and is exposed precisely because a heterogeneous dual-judge protocol was used. P1 therefore provides the upper-bound reference for personalisation quality against which the privacy-preserving P3 condition is evaluated in the next section.

## *4.4 LaaJ Evaluation of P3 (Paraphrase tweet based onconverted profile)*

The third evaluation condition (P3) examines the personalisation behaviour of the pipeline when the user profile supplied to the generator is the converted rather than the original profile used in P1. Whereas P1 establishes an upper bound on personalisation quality by exposing the generator to the user's unmodified writing history, P3 is designed to measure how much of that personalisation signal survives once the five anonymisation constraints have been applied to the profile. The comparison between P1 and P3 is therefore the central locus of the privacy/utility trade-off that motivates this study: any reduction in stylistic fidelity observed under P3 can be attributed specifically to the suppression of personally identifying information from the conditioning context.

*Table 3 Pairwise evaluation results of P3.*

| Judge Model | GT Selected | GT Rate | Model Selected | Model Rate | Total |
|---|---|---|---|---|---|
| Gemini 2.0 Flash | 194 | 77.60% | 56 | 22.40% | 250 |
| OpenAI GPT-4.1 | 241 | 96.40% | 9 | 3.60% | 250 |
| Average | 217.5 | 87.00% | 32.5 | 13.00% | 250 |

*GT = human-authored ground truth; Model = Claude-generated paraphrase. Each judge evaluated 250 samples from the LaMP-7 dataset.*

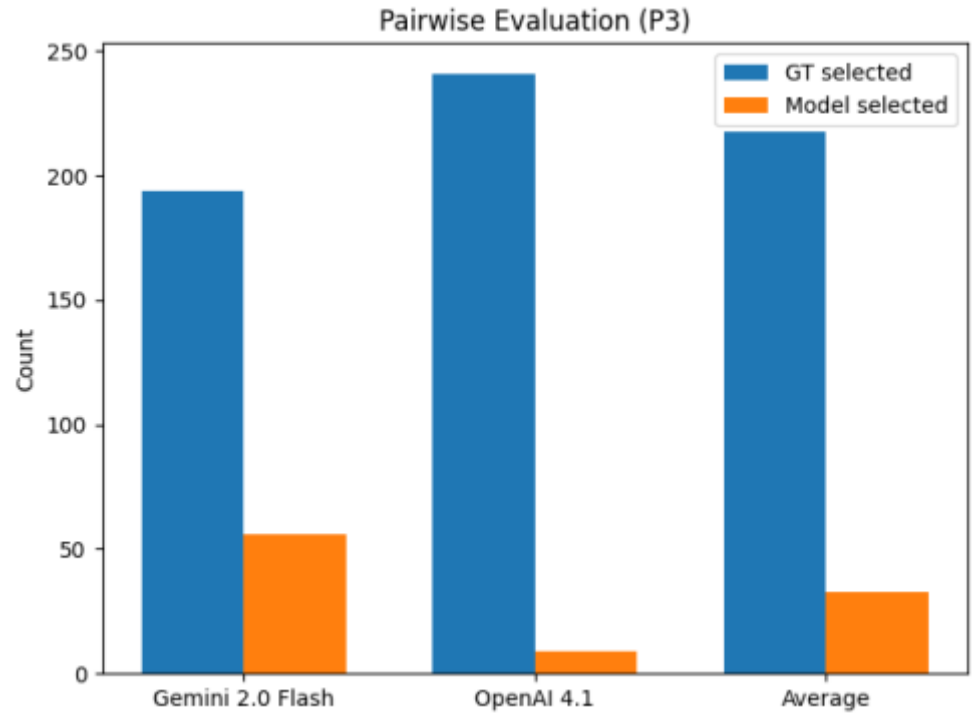

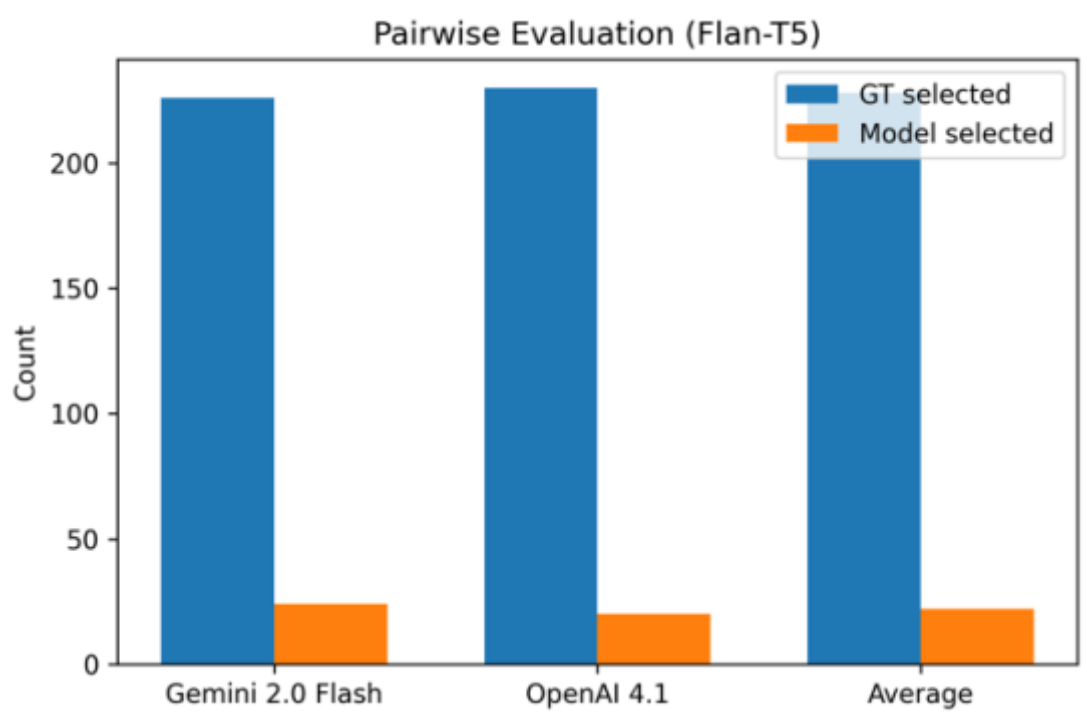

*Figure 18 Pairwise evaluation results for P3: (a) Claude Sonnet 4, (b) Google FlanT5.*

As reported in Table 3 and illustrated in Figure 18, both judges exhibited a pronounced preference for the ground-truth tweets over the paraphrased variants. Gemini 2.0 Flash selected the GT in 77.60% of cases (194/250), while OpenAI GPT-4.1 did so in 96.40% of cases (241/250), yielding an average GT selection rate of 87.00% across judges. Correspondingly, the paraphrased outputs were selected in only 13.00% of cases on average, substantially below the 50% threshold that would indicate stylistic indistinguishability between the two sources.

These results indicate that, while the rewritten text keeps the same meaning, it loses the user's unique personalization. The model is less able to reproduce the distinctive tone, writing style, and potentially identifiable linguistic features of the original author.

### *4.4.1 Inter-Judge Agreement Analysis*

To examine how the two judges arrived at the aggregate P3 result and whether their preferences were grounded in the same items, we conducted the same agreement analysis as for P1 on the 250 samples of the converted-profile condition. The results are presented in Table *4* and visualised in Figure 19. From the Figure 15 and Figure 18 we can see that FlanT5 generation is not distinguishable and it is not properly paraphrsed. So, here we are considering only claude generated tweet results. Hence, in the following figures and table we are presenting only Claude Sonnet generated tweet paraphrased results.

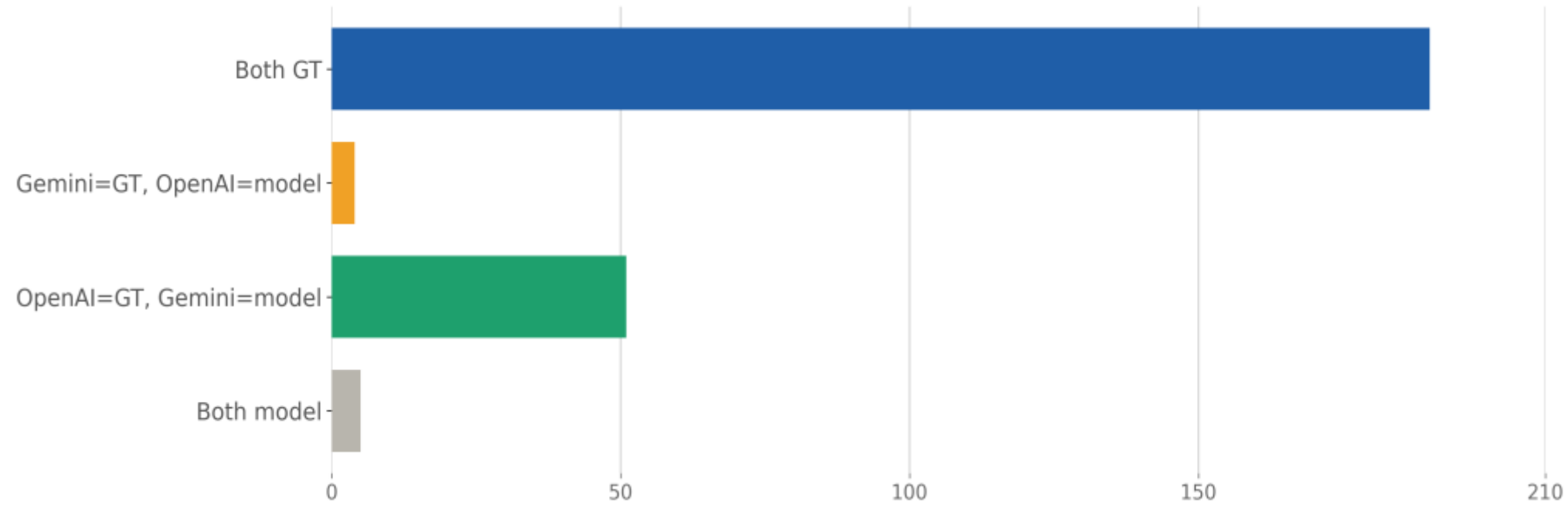

Figure 19 Cross-judge agreement analysis for P3

Table 4 Inter-judge agreement analysis for P3 (OpenAI 4.1 vs. Gemini 2.0 Flash, n = 250)

| Category | Count | % of Total |
|---|---|---|
| Both judges agree | 193 | 77.2% |
| Both select GT | 189 | 75.6% |
| Both select Model | 4 | 1.6% |
| Judges disagree | 57 | 22.8% |
| Gemini = GT, OpenAI = Model | 5 | 2.0% |
| Gemini = Model, OpenAI = GT | 52 | 20.8% |
| Total | 250 | 100.0% |

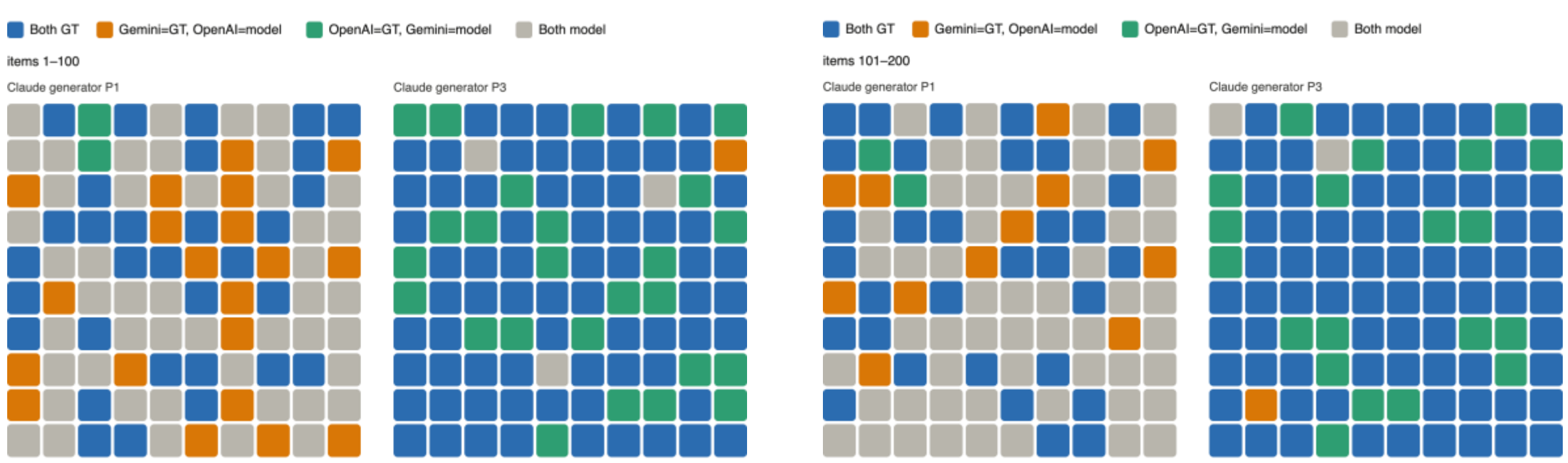

*Figure 20 Per-item agreement heatmap Claude (P1 vs P3)*

The inter-judge agreement analysis for P3 reveals a strong and consistent preference for the human-authored ground truth across both judges, as shown in Figure 19 and Figure 20. Both judges agreed on 193 out of 250 cases (77.2%), with the dominant agreement pattern being both judges simultaneously selecting the ground truth, occurring in 189 cases (75.6%). In contrast, both judges preferring the model-generated paraphrase was observed in only 4 cases (1.6%), indicating that the anonymised profile condition rarely produced outputs that either judge found more stylistically representative of the original author than the human-authored reference. Among the 57 disagreement cases (22.8%), the pattern was markedly asymmetric: in 52 cases (20.8%), OpenAI GPT-4.1 selected the ground truth while Gemini 2.0 Flash preferred the model output, whereas the reverse occurred in only 5 cases (2.0%), suggesting that GPT-4.1 consistently applies a stricter standard of stylistic fidelity. The per-item agreement heatmap in Figure 20 provides a complementary item-level view of this behaviour: comparing the P1 and P3 panels side by side. The P3 condition exhibits a visibly higher density of blue cells indicating both judges selecting the ground truth. Alongside a marked reduction in orange and grey cells relative to P1.

From a privacy perspective, this outcome is desirable. It implies that the paraphrasing pipeline effectively removes or obfuscates personal and identifying signals embedded in the original text. At the same time, this suppression of stylistic cues limits the model's ability to generate outputs that are perceived as authentic continuations of the original profile. Overall, the findings highlight a trade-off between privacy preservation and stylistic fidelity, where stronger anonymization leads to reduced personalization in the generated paraphrases.

### *4.5 Human Evaluation*

We conducted a human evaluation using a Google Form with 15 respondents. Evaluators were presented with a reference Twitter conversation and asked to judge which of two candidate responses sounded more like the original speaker. Each evaluator was a post-graduate student, and before they responded the study was explained to them. Two experimental conditions were tested. In Condition 1, evaluators compared an LLM-generated response produced from the original (unmodified) conversation against the ground-truth reply. In Condition 2, evaluators compared the ground-truth reply against an LLM-generated response produced from a paraphrased version of the conversation, where demographic identifiers, pop-culture references, personal details, and informal linguistic markers (slang, abbreviations, filler words) had been neutralized according to a standardized prompt. Evaluators could select one of four options: A, B, Both, or Neither.

*Table 5 Human evaluation results (n = 15) comparing LLM-generated responses against ground-truth replies under two conditions.*

| Condition | Input to LLM | A (%) | B (%) | Both (%) | Neither (%) |
|---|---|---|---|---|---|
| 1 | Original conversation | 33.3 (LLM) | 46.7 (Ground truth) | 20.0 | 0.0 |
| 2 | Paraphrased conversation | 93.3 (Ground truth) | 0.0 (LLM) | 6.7 | 0.0 |

Ⓐ "@thatswhack74 aww that's so sweet! i ended up makin a lil diy card n wrote em a song. nothin super fancy or whatever but it's all good" Ⓑ "@t...on sounds more like the person in the conversation?
15 responses

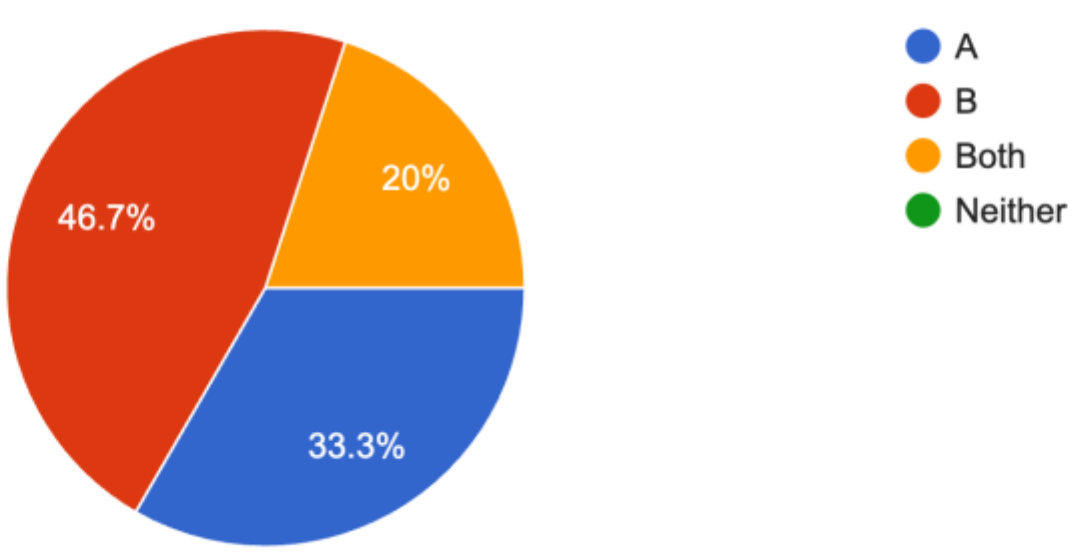

*Figure 21 Human evaluation results (n = 15) for Condition 1: evaluators compared an LLM-generated response (A) against the ground-truth reply (B), both conditioned on the original Twitter conversation.*

Ⓐ "@thatswhack74 aww that's sweet! i made a home made card and i wrote her a song... it wasn't that much but oh well.." Ⓑ "@thatswhack74 That ...n sounds more like the person in the conversation?
15 responses

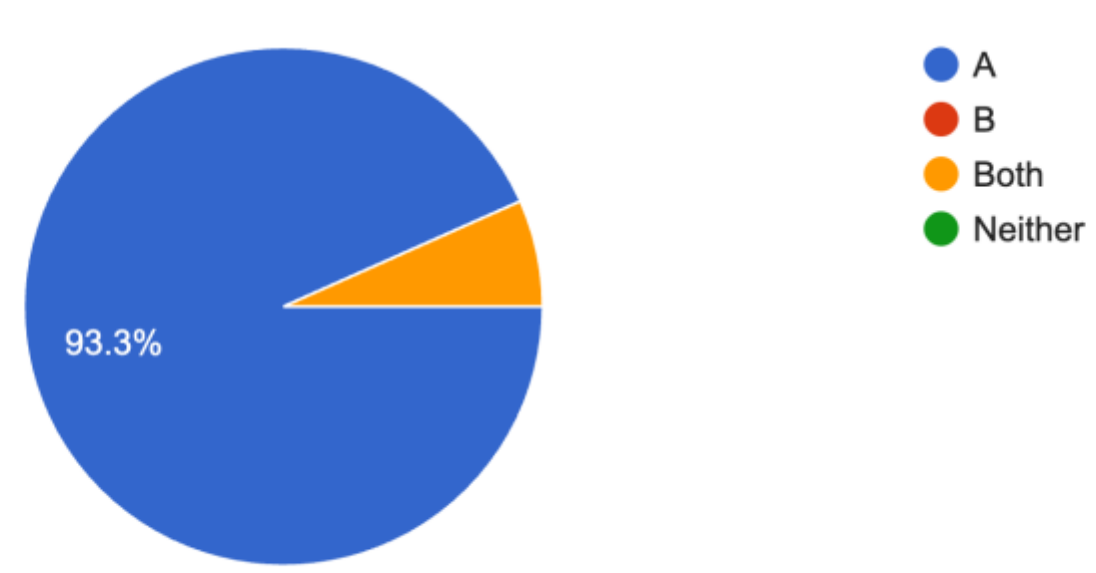

*Figure 22 Human evaluation results (n = 15) for Condition 2: evaluators compared the ground-truth reply (A) against an LLM-generated response (B) conditioned on the paraphrased conversation, in which demographic and informal stylistic cues had been neutralized.*

The results, summarized in Table 5, reveal a pronounced effect of the paraphrasing step on perceived authenticity. In Condition 1, the ground-truth response was preferred by 46.7% of evaluators, while the LLM-generated response was selected by 33.3%, with an additional 20.0% judging both responses equally representative of the speaker. This relatively close distribution indicates that, when stylistic cues are preserved in the input, the LLM can generate responses that approximate the speaker's voice with reasonable fidelity. In contrast, Condition 2 showed a strong preference for the ground-truth response (93.3%), with no evaluator selecting the LLM-generated response as more authentic and only 6.7% rating both as equivalent.

This sharp asymmetry suggests that the paraphrasing procedure removes stylistic signals critical to capturing a speaker's voice, and that this information loss cannot be recovered by the downstream generation model. These findings indicate that, LLMs are capable of producing persona-consistent text when provided with stylistically rich input. However, aggressive neutralization of demographic and informal cues substantially degrades the authenticity of generated responses.

### *4.6 Discussion*

The evaluation results provide clear evidence of a trade-off between personalization and privacy in LLM-generated text. When conditioned on original user profiles, the model produces paraphrases that are nearly indistinguishable from human-authored text. However, after profile anonymization, the quality of personalization drops significantly, with both LLM-based and human evaluations showing a strong preference for ground-truth outputs. Notably, semantic content remains largely preserved despite this decline. It proves that anonymization primarily affects stylistic signals rather than meaning. These findings show that the features that help models personalize text are the same ones that can reveal a user's identity. That is the feature creating a trade-off between protecting privacy and producing personalized outputs.

## 5 Conclusion

This study investigates whether unavoidable personal information affects personalized text generation. It showed that when LLMs are conditioned on original user tweet histories, they can generate closely personalized outputs. The profiles were then converted through profile-level paraphrasing designed to suppress demographic, personal, and stylistically identifying cues. The results show that this transformation preserves contextual meaning while successfully anonymizing the profile. Both LLM-based and human evaluations indicate that anonymization weakens the model's ability to reproduce the original speaker's voice. This study demonstrates the strength of current personalization methods and highlights an important privacy risk: historical user text contains stylistic patterns that can be reproduced convincingly by downstream models.

Overall, the findings demonstrate a fundamental trade-off between privacy preservation and stylistic fidelity in LLM personalization. Stronger anonymization reduces the risk of identity leakage but also limits the model's ability to generate highly personalized text. Future work should explore finer-grained privacy controls that selectively suppress identifying traits while preserving task-relevant style and meaning.

### DECLARATION

OpenAI GPT-5 and Claude Sonnet 4 were used to help express the original ideas in written form and to format the writing style for the academic paper.

### DATA AVALILABILITY

The source code, prompts, evaluation scripts, and experimental configuration used in this study are publicly available through Zenodo at https://doi.org/10.5281/zenodo.20934407 and https://sites.google.com/view/privacy-personalization-llms/home. The archived repository corresponds to the GitHub release associated with this manuscript and includes the materials necessary to reproduce the reported experiments. The LaMP7 dataset can be obtained from its original source as described in Reference [2]. The repository includes documentation to facilitate reproduction of the reported experiments.

### REFERENCES

[1] Patrick Lewis, Ethan Perez, Aleksandra Piktus, Fabio Petroni, Vladimir Karpukhin, Naman Goyal, Heinrich Küttler, Mike Lewis, Wen-tau Yih, Tim Rocktäschel, Sebastian Riedel, and Douwe Kiela. 2020. Retrieval-augmented generation for knowledge-intensive NLP tasks. In *Proceedings of the 34th International Conference on Neural Information Processing Systems (NeurIPS '20)*. Curran Associates Inc., Red Hook, NY, USA, Article 793, 9459–9474. https://doi.org/10.5555/3495724.3496517

[2] Alireza Salemi, Sheshera Mysore, Michael Bendersky, and Hamed Zamani. 2024. LaMP: When large language models meet personalization. In *Proceedings of the 62nd Annual Meeting of the Association for Computational Linguistics (Volume 1: Long Papers)*. Association for Computational Linguistics, 7370–7392. https://doi.org/10.18653/v1/2024.acl-long.399

[3] Michael Brennan, Sadia Afroz, and Rachel Greenstadt. 2012. Adversarial stylometry: Circumventing authorship recognition to preserve privacy and anonymity. *ACM Trans. Inf. Syst. Secur.* 15, 3 (Nov. 2012), Article 12. https://doi.org/10.1145/2382448.2382450

[4] Arvind Narayanan, Hristo Paskov, Neil Zhenqiang Gong, John Bethencourt, Emil Stefanov, Eui Chul Richard Shin, and Dawn Song. 2012. On the feasibility of internet-scale author identification. In *Proceedings of the 2012 IEEE Symposium on Security and Privacy*. IEEE, 300–314. https://doi.org/10.1109/SP.2012.46

[5] Lin Zhang and Hongyu Zhang. 2026. De-anonymization at scale via tournament-style attribution. arXiv:2601.12407. Retrieved from http://arxiv.org/abs/2601.12407

[6] Lianmin Zheng, Wei-Lin Chiang, Ying Sheng, Siyuan Zhuang, Zhanghao Wu, Yonghao Zhuang, Zi Lin, Zhuohan Li, Dacheng Li, Eric P. Xing, Hao Zhang, Joseph E. Gonzalez, and Ion Stoica. 2023. Judging LLM-as-a-judge with MT-Bench and Chatbot Arena. In *Proceedings of the 37th International Conference on Neural Information Processing Systems (NeurIPS '23)*. Curran Associates Inc., Red Hook, NY, USA, Article 2020, 46595–46623.

[7] Alireza Salemi, Surya Kallumadi, and Hamed Zamani. 2024. Optimization methods for personalizing large language models through retrieval augmentation. In *Proceedings of the 47th International ACM SIGIR Conference on Research and Development in Information Retrieval (SIGIR '24)*. Association for Computing Machinery, New York, NY, USA, 752–762. https://doi.org/10.1145/3626772.3657783

[8] Sheshera Mysore, Zhuoran Lu, Mengting Wan, Longqi Yang, Steve Menezes, Tina Baghaee, Emmanuel Barajas Gonzalez, Jennifer Neville, and Tara Safavi. 2024. PEARL: Personalizing large language model writing assistants with generation-calibrated retrievers. In *Proceedings of the 1st Workshop on Customizable NLP (CustomNLP4U)*. Association for Computational Linguistics, 198–219.

[9] Alireza Salemi and Hamed Zamani. 2024. Comparing retrieval-augmentation and parameter-efficient fine-tuning for privacy-preserving personalization of large language models. arXiv:2409.09510. Retrieved from https://doi.org/10.48550/arXiv.2409.09510

[10] Zhaoxuan Tan, Qingkai Zeng, Yijun Tian, Zheyuan Liu, Bing Yin, and Meng Jiang. 2024. Democratizing large language models via personalized parameter-efficient fine-tuning. In *Proceedings of the 2024 Conference on Empirical Methods in Natural Language Processing (EMNLP 2024)*. Association for Computational Linguistics, 6476–6491.

[11] Arta Misini, Arbana Kadriu, and Ercan Canhasi. 2022. A survey on authorship analysis tasks and techniques. *SEEU Review* 17, 2 (2022), 153–167. https://doi.org/10.2478/seeur-2022-0100

[12] Abiodun Modupe, Turgay Celik, Vukosi Marivate, and Oludayo O. Olugbara. 2022. Post-authorship attribution using regularized deep neural network. *Applied Sciences* 12, 15 (2022), 7518. https://doi.org/10.3390/app12157518

[13] Javier Huertas-Tato, Alejandro Martin, and David Camacho. 2025. PART: Pre-trained authorship representation transformer. *Human-centric Computing and Information Sciences* 14. Retrieved from http://arxiv.org/abs/2209.15373

[14] Malik H. Altakrori, Jackie C. K. Cheung, and Benjamin C. M. Fung. 2021. The topic confusion task: A novel evaluation scenario for authorship attribution. In *Findings of the Association for Computational Linguistics: EMNLP 2021*. Association for Computational Linguistics, 4242–4256. https://doi.org/10.18653/v1/2021.findings-emnlp.359

[15] Nils Lukas, Ahmed Salem, Robert Sim, Shruti Tople, Lukas Wutschitz, and Santiago Zanella-Béguelin. 2023. Analyzing leakage of personally identifiable information in language models. In *Proceedings of the 2023 IEEE Symposium on Security and Privacy (SP)*. IEEE, 346–363. https://doi.org/10.1109/SP46215.2023.10179300

[16] Siwon Kim, Sangdoo Yun, Hwaran Lee, Martin Pae Pilanavong, and Sungroh Yoon. 2023. ProPILE: Probing privacy leakage in large language models. In *Advances in Neural Information Processing Systems 36 (NeurIPS 2023)*.

[17] Badhan Chandra Das, M. Hadi Amini, and Yanzhao Wu. 2025. Security and privacy challenges of large language models: A survey. *ACM Comput. Surv.* 57, 6, Article 152 (June 2025). https://doi.org/10.1145/3712001

[18] Xiongtao Sun, Gan Liu, Zhipeng He, Hui Li, and Xiaoguang Li. 2024. DePrompt: Desensitization and evaluation of personal identifiable information in large language model prompts. arXiv:2408.08930. Retrieved from https://arxiv.org/abs/2408.08930

[19] Yijia Xiao, Yiqiao Jin, Yushi Bai, Yue Wu, Xianjun Yang, Xiao Luo, Wenchao Yu, Xujiang Zhao, Yanchi Liu, Quanquan Gu, Haifeng Chen, Wei Wang, and Wei Cheng. 2024. Large language models can be contextual privacy protection learners. In *Proceedings of the 2024 Conference on Empirical Methods in Natural Language Processing (EMNLP 2024)*. Association for Computational Linguistics, 14179–14201.

[20] Arjun Panickssery, Samuel R. Bowman, and Shi Feng. 2024. LLM evaluators recognize and favor their own generations. In *Advances in Neural Information Processing Systems 37 (NeurIPS 2024)*.

[21] Koki Wataoka, Tsubasa Takahashi, and Ryokan Ri. 2025. Self-preference bias in LLM-as-a-judge. arXiv:2410.21819. Retrieved from https://arxiv.org/abs/2410.21819

[22] Ryan Koo, Minhwa Lee, Vipul Raheja, Jong Inn Park, Zae Myung Kim, and Dongyeop Kang. 2024. Benchmarking cognitive biases in large language models as evaluators. In *Findings of the Association for Computational Linguistics: ACL 2024*. Association for Computational Linguistics, 517–545.

[23] Yijiang River Dong, Tiancheng Hu, and Nigel Collier. 2024. Can LLM be a personalized judge? In *Findings of the Association for Computational Linguistics: EMNLP 2024*. Association for Computational Linguistics, 10126–10141.

[24] Reza Staab, Robin Staab, Marc Langheinrich, and Martin Vechev. 2024. Beyond memorization: Violating privacy via inference with LLMs. In *Proceedings of the International Conference on Learning Representations (ICLR '24)*.

[25] Constantinos Patsakis and Nikolaos Lykousas. 2023. Man vs the machine in text anonymisation. Scientific Reports 13, 1 (2023).

[26] Zhuoran Li, Bin Guo, and Zhiwen Yu. 2026. Privacy control in conversational LLM platforms. In Proceedings of the 2026 CHI Conference on Human Factors in Computing Systems (CHI '26). Association for Computing Machinery, New York, NY, USA.

[27] Yifan Du, Zican Dong, Jinpeng Wang, Chuhan Wu, Wayne Xin Zhao, and Ji-Rong Wen. 2025. Automated profile inference with language model agents. arXiv:2505.12402. Retrieved from https://arxiv.org/abs/2505.12402

[28] Bingyang Zhang and Yulai Zhang. 2026. Assessing deanonymization risks with stylometry-assisted LLM agent. arXiv:2602.23079. Retrieved from https://arxiv.org/abs/2602.23079

[29] Nicole Mariah Sharon Belvisi, Naveed Muhammad, and Fernando Alonso-Fernandez. 2020. Forensic Authorship Analysis of Microblogging Texts Using N-Grams and Stylometric Features. In Proceedings of the 2020 IEEE 8th International Workshop on Biometrics and Forensics (IWBF '20). IEEE, New York, NY, USA.

[30] Chin-Yew Lin. 2004. ROUGE: A Package for Automatic Evaluation of Summaries. In Proceedings of the ACL Workshop on Text Summarization Branches Out. Association for Computational Linguistics, Barcelona, Spain, 74–81.

[31] Satanjeev Banerjee and Alon Lavie. 2005. METEOR: An Automatic Metric for MT Evaluation with Improved Correlation with Human Judgments. In Proceedings of the ACL Workshop on Intrinsic and Extrinsic Evaluation Measures for Machine Translation and/or Summarization. Association for Computational Linguistics, Ann Arbor, Michigan, USA, 65–72.

## APPENDIX

*Table 6 Sample flagging table from the dataset*

| id | f1 | f2 | f3 | f4 | f5 | #Flags |
|---|---|---|---|---|---|---|
| 610 | 0 | 1 | 1 | 1 | 1 | 4 |
| 6113 | 1 | 1 | 1 | 1 | 1 | 5 |
| 61152 | 0 | 1 | 1 | 1 | 1 | 4 |
| 61180 | 0 | 0 | 1 | 1 | 1 | 3 |
| 61188 | 0 | 1 | 1 | 0 | 1 | 3 |
| 61209 | 0 | 1 | 1 | 1 | 1 | 4 |
| 61210 | 0 | 0 | 1 | 1 | 1 | 3 |
| 61213 | 0 | 0 | 1 | 1 | 1 | 3 |
| 61249 | 0 | 1 | 1 | 1 | 1 | 4 |
| 6125 | 0 | 1 | 1 | 1 | 1 | 4 |
| 615 | 0 | 0 | 1 | 1 | 1 | 3 |
| 6150 | 1 | 1 | 1 | 1 | 1 | 5 |
| 6160 | 0 | 1 | 1 | 1 | 1 | 4 |
| 6171 | 0 | 0 | 1 | 1 | 1 | 3 |
| 6177 | 1 | 1 | 0 | 1 | 1 | 4 |

*Table 7 Profile risk scoring and risk level in sample profiles*

| id | f1 | f2 | f3 | f4 | f5 | #Flags | Score | Risk Level |
|---|---|---|---|---|---|---|---|---|
| 610 | 0 | 1 | 1 | 1 | 1 | 4 | 0.9333 | High |
| 6113 | 1 | 1 | 1 | 1 | 1 | 5 | 1.0000 | High |
| 61152 | 0 | 1 | 1 | 1 | 1 | 4 | 0.9333 | High |
| 61180 | 0 | 0 | 1 | 1 | 1 | 3 | 0.8000 | High |
| 61188 | 0 | 1 | 1 | 0 | 1 | 3 | 0.7333 | High |
| 61209 | 0 | 1 | 1 | 1 | 1 | 4 | 0.9333 | High |
| 61210 | 0 | 0 | 1 | 1 | 1 | 3 | 0.8000 | High |
| 61213 | 0 | 0 | 1 | 1 | 1 | 3 | 0.8000 | High |
| 61249 | 0 | 1 | 1 | 1 | 1 | 4 | 0.9333 | High |
| 6125 | 0 | 1 | 1 | 1 | 1 | 4 | 0.9333 | High |
| 615 | 0 | 0 | 1 | 1 | 1 | 3 | 0.8000 | High |
| 6150 | 1 | 1 | 1 | 1 | 1 | 5 | 1.0000 | High |
| 6160 | 0 | 1 | 1 | 1 | 1 | 4 | 0.9333 | High |
| 6171 | 0 | 0 | 1 | 1 | 1 | 3 | 0.8000 | High |
| 6177 | 1 | 1 | 0 | 1 | 1 | 4 | 0.7333 | High |

*Table 8 OpenAI evaluation response samples of P1*

<table>
<tr><th>id</th><th>paraphrased</th><th>ground_truth</th><th>openai_evaluation</th><th>Selection</th></tr>
<tr><td>610</td><td>i have loads to do today, but i'm so bored right now</td><td>got a busy day today.. but i'm bored right now</td><td>| Selected Paraphrase | Justification |<br>|--------------------|--------------|<br>| GT | The GT paraphrase ("got a busy day today.. but i'm bored right now") is more aligned with the user's conversational style and personalized tone. The profile demonstrates a preference for relaxed, slightly informal, and truncated sentence structures with ellipses and casual language (e.g., "yaa", "haha", "goshhh", use of "..", contractions, and omitting "I" at the start). GT's use of "got a busy day" and the double dots ".." reflects both the informality and brevity observed across the profile, making it sound much more like how this user writes or tweets compared to Para's full-sentence, standard phrasing with "I have loads to do today, but I'm so bored right now." GT better captures the user's natural voice. |</td><td>model</td></tr>
<tr><td>611</td><td>Wish Torchwood Children of Earth was on now.</td><td>I really wish Torchwood Children of Earth was on now</td><td>| Selection | Justification<br>| **GT** | The ground truth paraphrase ("I really wish Torchwood Children of Earth was on now") is more personalised, as the user frequently uses first-person expressions ("I", "I've", "I'm") and shows strong personal sentiments and conversational tone in their tweets. The inclusion of "really" and "I" mirrors their typical style and makes it more in line with their online persona. The generated paraphrase lacks the explicit "I" and emotional emphasis, making it less personal for this user. |</td><td>GT</td></tr>
<tr><td>612</td><td>Walkin' to my Anthro class & loving this cool weather. California weather is the BOMB!!!</td><td>Enjoying the cool weather as I walk to my Anthro class I love the cali weather...</td><td>**Selection: GT (Ground Truth)**<br>**Justification:**<br>The GT paraphrase is more personalised for this profile. The user frequently uses casual language, expressions of enjoyment, and affectionate references to California ("cali weather..."). GT matches this tone and style by informally stating "I love the cali weather..." and placing emphasis on personal enjoyment while walking to class. In contrast, the PARA version uses the less personalised "California weather is the BOMB!!!" which feels less aligned with the user's more relaxed and affectionate style based on the examples in the profile.</td><td>model</td></tr>
<tr><td>613</td><td>@ItsChelseaStaub Great!! Can't wait to see it!! But since I'm in Australia, I'll have to wait until next month to watch it!</td><td>@ItsChelseaStaub Awesome!! Can't wait to watch it!! Except I live in Australia so we won't be able to see it until next month!</td><td>| Selection | Justification |<br>|-----------|---------------|<br>| GT | The Ground Truth (GT) paraphrase is more personalized for this profile because the profile mentions being in Australia ("Nicole from Perth, Australia", references to "JonasAustralia"), and the GT version explicitly says "Except I live in Australia so we won't be able to see it until next month!"—using "we" to emphasize group identity and directly referencing location. In contrast, the model paraphrase uses "But since I'm in Australia," which is less conversational and less personalized. The GT feels more natural and relatable to this profile that often brings up their Australian identity in interactions.</td><td>GT</td></tr>
<tr><td>614</td><td>@Uheard My fav stores have generic laundry, dish soaps & fab softeners - I love my clothesline!</td><td>@Uheard Yes, I have noticed some stores have generics on laundry, dishwasher soaps & fabric softners - Now I'm in love w/my clothesline</td><td>**Selection:** GT<br>**Justification:** The ground truth paraphrase (GT) reflects more personalization for the profile because it includes interactive language ("Yes, I have noticed..."), and observes a personal journey ("Now I'm in love w/my clothesline"). This matches the style in the profile, which frequently uses conversational, affirming phrases, and shows self-disclosure and evolving habits ("I was out of town w/family stuff...," "I love all peppers actually..."). The generated para, by contrast, is more general and lacks the "Yes, I have noticed..." engagement as well as the explicit mention of change/growth, both of which are common in the profile's posts.</td><td>model</td></tr>
</table>

*Table 9 Gemini evaluation response samples of P3*

<table>
<tr><th>id</th><th>paraphrased</th><th>ground_truth</th><th>Gemini_evaluation</th><th>Selection</th></tr>
<tr><td>610</td><td>Got loads to do today but just sitting here doing nothing lol</td><td>got a busy day today.. but i'm bored right now</td><td>Here's the breakdown:<br><br>| Selection | Justification<br>|<br>| :-------- | :-------------------------------------------------------------------------------------------------------- |<br>| Para | The profile shows someone who uses "lol" and abbreviations. The generated paraphrase captures this better. |</td><td>model</td></tr>
<tr><td>611</td><td>Wish Torchwood Children of Earth was still on air right now.</td><td>I really wish Torchwood Children of Earth was on now</td><td>(Para/GT) GT<br><br>**Justification:**<br><br>The ground truth paraphrase uses "now" instead of "still on air right now", which aligns better with the user's informal and often concise communication style on Twitter. The user also frequently uses contractions and informal language in their tweets. The generated paraphrase is more formal and verbose.</td><td>model</td></tr>
<tr><td>612</td><td>Walking to Anthro class and this cool weather is just lovely. Gotta love California weather!</td><td>Enjoying the cool weather as I walk to my Anthro class I love the cali weather...</td><td>Here's the breakdown of which paraphrase is more personalized for the given profile:<br><br>| Selection | Justification |<br>|---|---|<br>| (GT) | The ground truth paraphrase includes "I love the cali weather," which aligns with the profile's generally enthusiastic and expressive tone. The use of "I" also is a more personalized sentiment, fitting the profile's style of sharing personal feelings. |</td><td>GT</td></tr>
<tr><td>613</td><td>@ItsChelseaStaub Awesome!! Can't wait to see it!! But being down in Australia means waiting till next month to catch it!</td><td>@ItsChelseaStaub Awesome!! Can't wait to watch it!! Except I live in Australia so we won't be able to see it until next month!</td><td>(GT) - The ground truth more explicitly mentions living in Australia, which is directly relevant to the user's previous tweets about Jonas Brothers Australia and asking for a shoutout from Perth, Australia.</td><td>GT</td></tr>
<tr><td>614</td><td>@Uheard noticed some stores have generic options for laundry, dish soaps and fabric softeners. Loving the clothesline choice - small steps toward change!</td><td>@Uheard Yes, I have noticed some stores have generics on laundry, dishwasher soaps & fabric softners - Now I'm in love w/my clothesline</td><td>Here's my assessment:<br><br>**(GT) |** The ground truth paraphrase captures the original sentiment and the specific items mentioned (laundry, dishwasher soaps & fabric softners). It also includes a personal touch ("Now I'm in love w/my clothesline"), aligning with the user's expressions of personal preferences and enjoyment of simple things.</td><td>GT</td></tr>
<tr><td>615</td><td>@AhmNoHere looks like there's some trouble going on there.</td><td>@AhmNoHere Sounds like a couple of "issues" there to me...</td><td>(Para/GT) | Justification<br>------- | --------<br>(GT) | The user uses informal language, abbreviations, and emojis, suggesting a more casual and less formal communication style. The ground truth reflects this better.</td><td>model</td></tr>
<tr><td>616</td><td>According to @QueenofScots67, Twitter will limit you if you go over 100 requests in an hour.</td><td>@QueenofScots67 if you make more than 100 "requests" in an hour you are stopped by twitter from using the system for a while. I think</td><td>(GT/GT)<br>**Justification:**<br>The user frequently interacts with others on Twitter, often in direct response to their tweets and in a conversational style. The Ground Truth paraphrase reflects this conversational tone and directly attributes the information to the original tweeter in a way that resembles a casual exchange. The Generated paraphrase, while grammatically correct and informative, lacks the personal touch and conversational feel that aligns with the user's typical Twitter interactions.</td><td>GT</td></tr>
</table>

*Table 10 Appendix examples of pairwise evaluation outputs for P1, showing model-generated paraphrases, ground truth references, and corresponding judge decisions from OpenAI 4.1 and Gemini 2.0 Flash, along with agreement outcomes.*

| ID | Paraphrase (Generated) | Ground Truth | OpenAI | Gemini | Agreement |
|---|---|---|---|---|---|
| 610 | i have loads to do today, but i'm so bored right now | got a busy day today.. but i'm bored right now | Para | Para | Agree |
| 611 | Wish Torchwood Children of Earth was on now. | I really wish Torchwood Children of Earth was on now | GT | GT | Agree |
| 612 | Walkin' to my Anthro class & loving this cool weather. California weather is the BOMB!!! | Enjoying the cool weather as I walk to my Anthro class I love the cali weather... | Para | Para | Agree |
| 613 | @ItsChelseaStaub Great!! Can't wait to see it!! But since I'm in Australia, I'll have to wait until next month to watch it! | @ItsChelseaStaub Awesome!! Can't wait to watch it!! Except I live in Australia so we won't be able to see it until next month! | GT | GT | Agree |
| 614 | @Uheard My fav stores have generic laundry, dish soaps & fab softeners - I love my clothesline! | @Uheard Yes, I have noticed some stores have generics on laundry, dish-washer soaps & fabric softeners - Now I'm in love w/my clothesline | Para | Para | Agree |
| 615 | @AhmNoHere, sounds like there's some issues goin' on over there. | @AhmNoHere Sounds like a couple of "issues" there to me... | GT | GT | Agree |
| 616 | @QueenofScots67 apparently you can only look at 100 things an hour on here now ... who knew ? | @QueenofScots67 if you make more than 100 "requests" in an hour you are stopped by twitter from using the system for a while. I think | Para | Para | Agree |
| 617 | @cyandle hey, anything with pie is a win, right? peach cobbler, cherry fried pies, banana cream... all solid gold. | @cyandle you name it, sug. if it ends in pie, it's a win course, peach (cobbler), cherry (fried pies), and banana cream top the list. | GT | GT | Agree |
| 6111 | Miss the Facebook hype about it three months ago? | @Patty_B You didn't see everyone going on about it on Facebook about 3 months ago? | Para | GT | Disagree |
| 6112 | @knitster good morning.. well, not for you, is it? | @knitster morning not morning for you though, is it? | GT | Para | Disagree |

*Table 11 Appendix examples of pairwise evaluation outputs for P3, showing model-generated paraphrases, ground truth references, and corresponding judge decisions from OpenAI 4.1 and Gemini 2.0 Flash, along with agreement outcomes.*

| ID | Paraphrase (Generated) | Ground Truth | OpenAI | Gemini | Agreement |
|---|---|---|---|---|---|
| 610 | Got loads to do today but just sitting here doing nothing lol | got a busy day today.. but i'm bored right now | GT | Para | Disagree |
| 611 | Wish Torchwood Children of Earth was still on air right now. | I really wish Torchwood Children of Earth was on now | GT | Para | Disagree |
| 612 | Walking to Anthro class and this cool weather is just lovely. Gotta love California weather! | Enjoying the cool weather as I walk to my Anthro class I love the cali weather... | GT | GT | Agree |
| 613 | @ItsChelseaStaub Awesome!! Can't wait to see it!! But being down in Australia means waiting till next month to catch it! | @ItsChelseaStaub Awesome!! Can't wait to watch it!! Except I live in Australia so we won't be able to see it until next month! | GT | GT | Agree |
| 614 | @Uheard noticed some stores have generic options for laundry, dish soaps and fabric softeners. Loving the clothesline choice - small steps toward change! | @Uheard Yes, I have noticed some stores have generics on laundry, dishwasher soaps & fabric softeners - Now I'm in love w/my clothesline | GT | GT | Agree |
| 615 | @AhmNoHere looks like there's some trouble going on there. | @AhmNoHere Sounds like a couple of "issues" there to me... | GT | Para | Disagree |
| 616 | According to @QueenofScots67, Twitter will limit you if you go over 100 requests in an hour. | @QueenofScots67 if you make more than 100 "requests" in an hour you are stopped by twitter from using the system for a while. I think | GT | GT | Agree |
| 617 | @cyandle honestly anything with pie in it is a win. Peach cobbler, cherry fried pies and banana cream are some of the best! | @cyandle you name it, sug. if it ends in pie, it's a win course, peach (cobbler), cherry (fried pies), and banana cream top the list. | Para | GT | Disagree |
| 6111 | Did you catch all the Facebook buzz about it back three months ago? | @Patty_B You didn't see everyone going on about it on Facebook about 3 months ago? | GT | GT | Agree |
| 6112 | @knitster Good morning! Well, it's morning for me at least - not so much for you I imagine? | @knitster morning not morning for you though, is it? | Para | Para | Agree |

*Table 12 Side-by-side comparison of automated and human evaluation results across conditions.*

| Metric | Original (P1 / Cond. 1) | Suppressed profile (P3 / Cond. 2) | Δ |
|---|---|---|---|
| Automated evaluation (LLM judges) | | | |
| Model preferred (avg.) | 49.69% | 13.00% | −36.69 pp |
| Ground truth preferred (avg.) | 50.31% | 87.00% | +36.69 pp |
| Both judges select Model | 26.1% | 1.6% | −24.5 pp |
| Both judges select GT | 26.9% | 75.6% | +48.7 pp |
| Inter-judge raw agreement | 53.0% | 77.2% | +24.2 pp |
| Cohen's κ | 0.078 | 0.065 | ≈ 0 |
| Disagreement asymmetry ratio | 1.85 : 1 | 10.4 : 1 | sharper |
| McNemar exact p | 0.0016 | $< 10^{-10}$ | stronger |
| Human evaluation (n = 15) | | | |
| Model preferred (strict) | 33.3% | 0.0% | −33.3 pp |
| Ground truth preferred (strict) | 46.7% | 93.3% | +46.6 pp |
| Both equivalent | 20.0% | 6.7% | −13.3 pp |
| Model at least equivalent | 53.3% | 6.7% | −46.6 pp |

*Table 13 System prompt for converting original conversation*

*You are a text paraphrasing assistant. Your task is to paraphrase the given tweet while following these strict rules:*

*1. Remove age and demographic identifiers --- Do not include any mention of the user's age or demographic group.*

*2. Generalise interests --- Replace specific pop culture references (e.g., celebrity names, movie franchises, TV shows) with neutral terms like 'musicians,' 'celebrities,' or 'entertainment.'*

*3. Redact personal life details --- Remove or generalise any references to family situations, personal events, or location-specific details.*

*4. Use generic language --- Avoid specific titles, names, or niche references. Replace them with broad, neutral equivalents.*

*5. Remove informal text and shorthand --- Convert casual language, slang, abbreviations, and filler words (e.g., 'lol', 'haha', 'aww', 'hun', 'hbu') into formal and structured language.*

*Return ONLY the paraphrased text with no explanation or preamble.*

*Table 14 The iterative audit-and-revise procedure of conversion process*

```
Algorithm: PII-Suppression with Adversarial Audit
Input:  user profile C_u^(0); generator G; auditor A;
        prompts π_conv, π_audit; max iterations T_max
Output: converted profile C_u*
1:  C ← G(π_conv || C_u^(0))         // initial paraphrasing pass
2:  F ← A(π_audit || C)              // first audit
3:  t ← 1
4:  while F ≠ ∅ and t < T_max do
5:      C ← G(π_conv || C || F)      // revise using auditor flags
6:      F ← A(π_audit || C)
7:      t ← t + 1
8:  end while
9:  return C_u* ← C
```

| Tweet to Paraphrase [610] | Phase 1 |
|---|---|
| Although I have a lot to do today, I am currently feeling unoccupied. | i have loads to do today, but i'm so bored right now |
| **Ground Truth** | **Phase 3 (After anonymization)** |
| got a busy day today.. but i'm bored right now | Got loads to do today but just sitting here doing nothing |

| Tweet to Paraphrase [615] | Phase 1 |
|---|---|
| @AhmNoHere, it seems like there are some problems there. | @AhmNoHere, sounds like there's some issues goin' on over there. |
| **Ground Truth** | **Phase 3 (After anonymization)** |
| @AhmNoHere Sounds like a couple of "issues" there to me... | @AhmNoHere Looks like there's some trouble going on there. |

*Figure 23 Phase 1 vs Phase 3 (sample comparison of Claude paraphrase)*